%% file: main.tex
\documentclass[10pt,twocolumn,letterpaper]{article}

\usepackage{iccv}
\usepackage{multirow}

\input{preamble}

\definecolor{iccvblue}{rgb}{0.21,0.49,0.74}
\usepackage[pagebackref,breaklinks,colorlinks,allcolors=iccvblue]{hyperref}

\def\paperID{}
\def\confName{ICCV}
\def\confYear{2025}

\title{HORT: Monocular Hand-held Objects Reconstruction with Transformers}

\author{Zerui Chen$^{1}$ \quad Rolandos Alexandros Potamias$^{2}$ \quad Shizhe Chen$^{1}$ \quad Cordelia Schmid$^{1}$ \vspace{0.12cm}\\
$^{1}$Inria, \'Ecole normale sup\'erieure, CNRS, PSL Research University \quad $^2$Imperial College London \\
{\tt\small firstname.lastname@inria.fr \quad r.potamias@imperial.ac.uk}\\
{\tt\small \url{https://zerchen.github.io/projects/hort.html}}\\
}

\begin{document}
\maketitle

\input{sec/0_abstract}    
\input{sec/1_intro}
\input{sec/2_related}
\input{sec/3_method}
\input{sec/4_expr}
\input{sec/5_conclusion}

{
    \small
    \bibliographystyle{ieeenat_fullname}
    \bibliography{main}
}

\clearpage
\section*{Appendix}
\input{appendix}

\end{document}

%% file: preamble.tex
\usepackage{xcolor}
\usepackage{multirow}
\usepackage[percent]{overpic}
\usepackage{graphicx}
\usepackage{pifont}
\usepackage{float}
\usepackage{cuted}
\usepackage[symbol]{footmisc}

%% file: sec/0_abstract.tex
\begin{abstract}
Reconstructing hand-held objects in 3D from monocular images remains a significant challenge in computer vision. 
Most existing approaches rely on implicit 3D representations, which produce overly smooth reconstructions and are time-consuming to generate explicit 3D shapes. While more recent methods directly reconstruct point clouds with diffusion models, the multi-step denoising makes high-resolution reconstruction inefficient.
To address these limitations, we propose a transformer-based model to efficiently reconstruct dense 3D point clouds of hand-held objects.
Our method follows a coarse-to-fine strategy, first generating a sparse point cloud from the image and progressively refining it into a dense representation using pixel-aligned image features.
To enhance reconstruction accuracy, we integrate image features with 3D hand geometry to jointly predict the object point cloud and its pose relative to the hand.
Our model is trained end-to-end for optimal performance.
Experimental results on both synthetic and real datasets demonstrate that our method achieves state-of-the-art accuracy with much faster inference speed, while generalizing well to in-the-wild images.
\end{abstract}

%% file: sec/1_intro.tex
\section{Introduction}
\label{sec:intro}
Reconstructing hand-held objects in 3D from monocular images has shown great potential for widespread real-world applications, ranging from action recognition~\cite{damen2018scaling,grauman2022ego4d,banerjee2024introducing,fan2024benchmarks}, human-computer interaction~\cite{chao2022handoversim,christen2023learning,christen2024synh2r} to robotics manipulation~\cite{mandikal2022dexvip,qin2022dexmv,chen2025vividex}. 
Despite notable advancements in this task, achieving robust and efficient 3D reconstruction remains a significant challenge.

\begin{figure}[t]
  \centering  
  \includegraphics[width=0.95\linewidth]{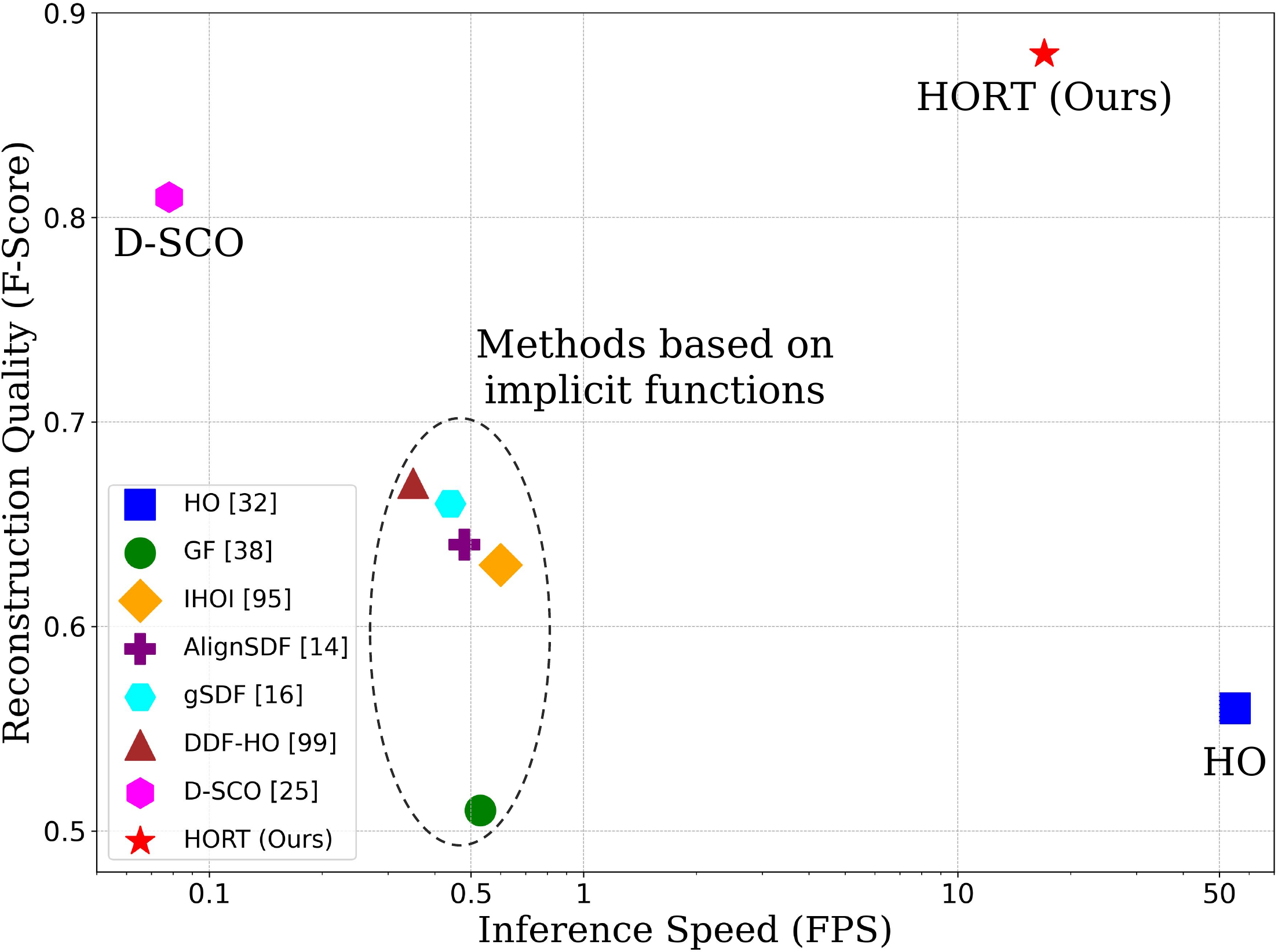}
  \vspace{-0.3cm}
  \caption{Comparison of inference speed and reconstruction quality on the ObMan dataset. Inference time includes meshing time. Implicit methods~\cite{karunratanakul2020grasping,ye2022s,chen2022alignsdf,chen2023gsdf} require approximately 2 seconds to generate the mesh but fail to achieve compelling results. Explicit methods face trade-offs between reconstruction quality and inference speed: HO~\cite{hasson2019learning} uses meshes while D-SCO~\cite{fu2024d} employs point clouds, but neither achieves optimal performance in both metrics. HORT demonstrates superior performance by achieving high-quality reconstruction with fast inference speed.}
  \label{fig:runtime}
  \vspace{-0.4cm}
\end{figure}

Existing approaches for hand-held object reconstruction can be broadly categorized into two types.
The first category employs optimization-based algorithms to predict 3D object poses~\cite{hampali2020honnotate,yang2021cpf,cao2021reconstructing,hasson2021towards,zhu2023get} alone or jointly with 3D shapes~\cite{huang2022reconstructing,ye2023diffusion,fan2024hold}. 
While these methods achieve promising reconstruction accuracy, they often rely on additional 3D inputs such as multi-view images, videos, depths or known object meshes. Moreover, they often require several hours to optimize the pose and shape using complex objective functions, making them impractical for real-time applications. 
To alleviate these issues, the second category explores learning-based approaches. A predominate line of work leverages deep implicit functions~\cite{park2019deepsdf,mescheder2019occupancy,chen2019learning,aumentado2022representing}, such as signed distance functions~\cite{karunratanakul2020grasping, chen2022alignsdf, ye2022s, chen2023gsdf, qi2024hoisdf} and directed distance functions~\cite{zhang2024ddf}, to reconstruct hand-held objects. However, methods based on implicit functions tend to generate overly smooth 3D surfaces, failing to capture fine geometric details~\cite{park2019deepsdf,choe2021deep}.
Additionally, obtaining explicit 3D meshes from learned implicit fields requires another post-processing step with the Marching Cubes algorithm~\cite{lorensen1987marching}, which not only reduces inference speed (Figure~\ref{fig:runtime}) but also limits flexibility for downstream tasks.

A few learning-based methods~\cite{hasson2019learning,fu2024d} have explored explicit 3D representations for hand-held object reconstruction, offering advantages for visualization and integration with downstream applications such as object CAD model registration and optimization-based pipeline~\cite{hasson2021towards,cao2021reconstructing,zhu2023get,wu2024reconstructing}. 
HO~\cite{hasson2019learning} employs vertex-based representation, achieving fast inference but suffering from limited reconstruction resolution.
Motivated by the recent success of point clouds in 3D reconstruction and generation~\cite{melas2023pc2,nichol2022point,kim2024multi}, the state-of-the-art method D-SCO~\cite{fu2024d} reconstructs high-resolution object point clouds from monocular images using diffusion models~\cite{song2020score}.
However, despite its strong performance, D-SCO is even slower than implicit function based methods, requiring over 13 seconds per reconstruction due to its intensive denoising steps at test time.

To balance inference speed and 3D reconstruction quality, we introduce HORT, a coarse-to-fine Transformer-based framework for Hand-held Object Reconstruction.~HORT consists of four key components: an image encoder, a hand encoder, a sparse point cloud decoder, and a dense point cloud decoder.
The two encoders extract complementary features - visual cues from the monocular image and geometric information from the hand.
To effectively encode the hand's kinematic structure, we transform the estimated hand vertices into multiple hand joint coordinate systems before processing them with a PointNet~\cite{qi2017pointnet} architecture. 
As hand pose and shape provide implicit cues about the object's geometry and position, our sparse point cloud decoder integrates both image and 3D hand features via a multi-layer transformer, and jointly predicts sparse object point clouds and hand-relative object pose. 
The dense point cloud decoder then upsamples these sparse representations to high resolution. For enhanced reconstruction fidelity, we project sparse points onto the image plane to retrieve pixel-aligned image features before applying the dense network for refinement. 
Our model is trained end-to-end with object reconstruction losses. 
We evaluate our model on both synthetic ObMan~\cite{hasson2019learning} dataset and real-world HO3D~\cite{hampali2020honnotate}, DexYCB~\cite{chao2021dexycb} and MOW~\cite{cao2021reconstructing} datasets. Extensive ablation studies validate the effectiveness of our approach. HORT achieves state-of-the-art reconstruction accuracy across all benchmarks while maintaining high inference speed.

Our contributions are summarized as follows:

\noindent
\textbf{$\bullet$} 
We propose a coarse-to-fine transformer-based framework (HORT) that efficiently reconstructs dense 3D point clouds of hand-held objects from monocular images, balancing reconstruction quality with computational efficiency.

\noindent
\textbf{$\bullet$} 
We integrate fine-grained image and hand geometry features for the sparse and dense point cloud generation, leveraging the implicit cues provided by hand shape for more accurate object reconstruction.

\noindent
\textbf{$\bullet$} Our approach outperforms state-of-the-art methods on four diverse benchmarks and generalizes well to in-the-wild images, while achieving fast inference speed.

%% file: sec/2_related.tex
\section{Related Work}
\label{sec:related}

\textbf{3D hand reconstruction.} 
We have witnessed remarkable progress in 3D hand reconstruction over the past decades, evolving from early efforts in the 1990s~\cite{heap1996towards,rehg1994visual} to modern deep learning approaches~\cite{lepetit2020recent_pose_advances,Yuan_2018_CVPR}.
Early deep learning based approaches~\cite{zimmermann2017learning,iqbal2018hand,tang2014latent,tekin2019h+,moon2018v2v,spurr2021self,xiong2019a2j,sun2018integral,mueller2018ganerated,meng20223d,chen2021camera,chen2020towards,chen2022learning} focus on predicting a sparse set of 3D hand joints from monocular images.
The introduction of the parametric MANO hand model~\cite{MANO:SIGGRAPHASIA:2017} has enabled more recent works~\cite{lv2021handtailor,wang2020rgb2hands,mueller2019real,kulon2019rec,Kulon2020weaklysupervisedmh,baek2019pushing,chen2021camera,boukhayma20193d,hampali2022keypoint,li2022interacting,pavlakos2024reconstructing,potamias2024wilor,rong2021frankmocap,lin2021end} to reconstruct dense 3D hand mesh with neural networks by integrating MANO as a differentiable layer.
More recently, state-of-the-art 3D hand reconstruction methods~\cite{potamias2024wilor,pavlakos2024reconstructing,zhang2025hawor} have shifted from convolutional neural networks to transformer networks~\cite{vaswani2017attention} and benefitted much from large-scale training, achieving robust 3D hand reconstruction and strong generalization to in-the-wild images. 
Our work focuses on hand-held object reconstruction, which is more challenging than hands due to the diverse nature of objects. We leverage high-quality automatic hand reconstruction as guidance to improve the accuracy of object reconstruction.

\begin{figure*}[ht]
  \centering
\includegraphics[width=0.95\linewidth]{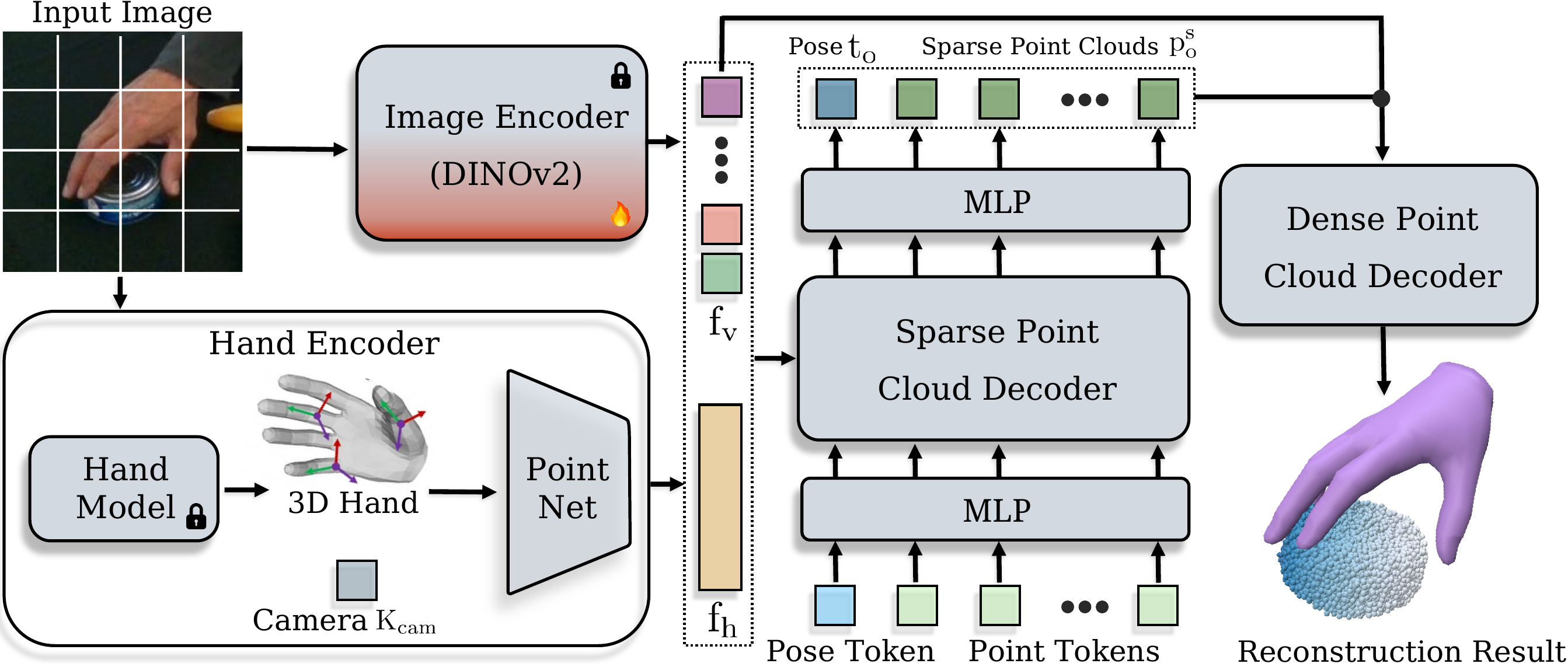}
  \vspace{-0.3cm}
  \caption{
  Overview of the proposed HORT model for 3D reconstruction of hand-held objects from a single RGB image. The model first extracts fine-grained visual and hand geometry features using an image encoder and a hand encoder. Then, the sparse and dense point cloud decoders integrate both visual and hand information to progressively generate object point clouds in a coarse-to-fine manner.
  }
  \vspace{-0.5cm}
  \label{fig:method}
\end{figure*}

\noindent \textbf{Hand-held object reconstruction via optimization.} 
The optimization-based method has been widely used to construct ground-truth 3D reconstructions of hand-held objects~\cite{chao2021dexycb,yang2022oakink,hampali2020honnotate,ballan2012motion,oikonomidis2011full,wang2013video,wang2024ho,wen2023bundlesdf}, typically leveraging extra 3D inputs such as multi-view images, depth sensors, or known 3D mesh models to produce accurate reconstructions. However, such inputs are typically unavailable in real-world applications.
To address this limitation, another line of work focuses on using optimization to refine initial predictions of object poses and shapes~\cite{hasson2021towards,cao2021reconstructing,yang2021cpf,zhu2023get,fan2024hold,ye2023diffusion,huang2022reconstructing,hampali2023hand,liu2025easyhoi}. These methods incorporate constraints to improve hand-object contact, prevent penetration, or enhance temporal consistency in monocular RGB videos. 
Some methods~\cite{hasson2021towards,cao2021reconstructing,yang2021cpf} optimize only 6D object poses given known CAD models with multiple objectives in 2D projection, hand-object interaction and collision. 
Other works~\cite{fan2024hold,ye2023diffusion,hampali2023hand} further relax the CAD assumption by jointly optimizing object poses and shapes parameterized by SDF or NeRF~\cite{mildenhall2020nerf}. A recent work EasyHOI~\cite{liu2025easyhoi} employs foundation models to reconstruct objects, followed by hand pose optimization to infer plausible hand and object configurations. It demonstrates impressive zero-shot performance on main benchmarks.~In general, optimization-based methods require complex objective functions and iterative refinement, making them prone to local minima and leading to slow inference speed.
In contrast, our method reconstructs hand-held objects in an efficient feed-forward manner. Our prediction results can also serve as strong initializations for accelerating optimization-based methods.

\noindent \textbf{Hand-held object reconstruction via learning.} 
Learning-based methods~\cite{hasson2019learning,tse2022collaborative,li2023chord,chen2022alignsdf,chen2023gsdf,ye2022s,zhang2024ddf,fu2024d,hu2024learning} train neural models to directly reconstruct hand-held objects from monocular images. 
Early works such as Hasson \etal~\cite{hasson2019learning} and Tse \etal~\cite{tse2022collaborative} employ AtlasNet~\cite{groueix2018papier} to reconstruct objects as sets of mesh vertices, which suffer from limited resolution and unrealistic geometry.
To improve detailed reconstruction, later works~\cite{karunratanakul2020grasping,chen2022alignsdf,chen2023gsdf,ye2022s,zhang2024ddf} adopt implicit 3D representations such as SDF. Despite the potential of continuous functions, these methods often generate overly smooth surfaces and are time-consuming to derive explicit 3D shapes. 
To tackle this issue, the state-of-the-art method D-SCO~\cite{ye2023diffusion} proposes to reconstruct explicit point clouds with diffusion models. However, the multi-step denoising process in diffusion models leads to low inference speed for high-resolution point clouds. 
In this work, we adopt explicit point clouds similar to D-SCO. However, our approach differs fundamentally by using a transformer-based architecture to reconstruct object point clouds in a coarse-to-fine manner, achieving state-of-the-art reconstruction accuracy while significantly improving inference speed.

%% file: sec/3_method.tex
\section{Method}
\label{sec:method}

In this section, we present the \textbf{HORT} model for \textbf{H}and-held \textbf{O}bject \textbf{R}econstruction based on \textbf{T}ransformers, which predicts dense object point clouds from monocular RGB images.
We first introduce the overall architecture in Section~\ref{sec:method_overview}, followed by a description of the detailed components in Section~\ref{sec:vfe} to ~\ref{sec:dense_recon}.
Finally, we describe the training objectives for end-to-end optimization of the HORT model in Section~\ref{sec:train}.

\subsection{Overview}
\label{sec:method_overview}

Figure~\ref{fig:method} illustrates the overview of our hand-held object reconstruction model HORT, which consists of four key modules: the image encoder, the hand encoder, the sparse point cloud decoder and the dense point cloud decoder.
We adopt a coarse-to-fine paradigm to progressively generate high-resolution point clouds for hand-held objects.

In the first stage, a sparse point cloud is generated using the image and hand encoders, along with the sparse point cloud decoder. 
The two encoders extract image features and 3D hand features, respectively. 
The image features provide crucial information about the object's appearance and 2D contours, while the 3D hand features help regularize the object's 3D shape and its spatial relationship to the hand, which enable to mitigate the inherent ambiguities in monocular RGB-based 3D reconstruction. 
Unlike previous approaches~\cite{ye2022s,chen2022alignsdf,chen2023gsdf} that only leverage sparse hand joint poses for object shape learning, our hand encoder utilizes fine-grained hand geometry, providing richer structural cues to infer the shape of manipulated objects. 
To effectively integrate the image and hand features, we introduce the sparse point cloud decoder, which jointly predicts the initial sparse point cloud and the object's pose relative to the hand.

In the second stage, we propose a dense point cloud decoder to refine and upsample the sparse point clouds generated in the first stage.
Leveraging the predicted camera parameters, we project the sparse point clouds onto the image plane, enabling to retrieve fine-grained, pixel-aligned image features for each sparse point.
These features provide rich local details to enhance the upsampling process, leading to high-resolution 3D object reconstruction.
This two-stage framework strikes a balance between reconstruction accuracy and computational efficiency. 

\subsection{Image and hand feature encoding}
\label{sec:vfe}

\noindent \textbf{Fine-grained hand features.}
Given a single monocular image as the input, we use a state-of-the-art hand pose estimation model~\cite{hasson2019learning,potamias2024wilor} to reconstruct the 3D hand and predict the camera parameters $\rm K_{cam}$, which includes a camera translation $\rm t_{c} \in \mathbb{R}^{3}$ and a scale factor $\rm s_{c} \in \mathbb{R}$ that normalizes the hand size in the camera coordinate system.
The estimated 3D hand mesh is parameterized using MANO~\cite{MANO:SIGGRAPHASIA:2017} and consists of 3D vertices ${\rm v_{h}} \in \mathbb{R}^{778\times3}$ driven by rigged hand joints ${\rm j_{h}} \in \mathbb{R}^{16\times3}$. To effectively encode the 3D hand, we transform ${\rm v_{h}}$ into multiple local coordinate systems including all hand joints, finger tips and palm, resulting in a total of $22$ coordinate systems ($16$ joints, $5$ fingertips, and $1$ palm). 
We further concatenate the transformed coordinates of each vertex with an absolute vertex index, which helps to better capture the spatial positioning of each vertex within the hand shape.
This yields a comprehensive hand representation ${\rm e_{h}} \in \mathbb{R}^{778\times(22\times3+1)}$ that captures both pose and shape dependent geometry.
We finally encode this rich geometric representation ${\rm e_{h}}$ through an efficient PointNet architecture~\cite{qi2017pointnet} to extract hand features ${\rm f_{h} \in \mathbb{R}^{1024}}$. 

\noindent \textbf{Image features.}
We employ DINOv2~\cite{oquab2023dinov2} to extract image features from the input image $\rm I \in \mathbb{R}^{224\times224\times3}$.
DINOv2 embeds images into 257 visual feature tokens ${\rm f_{v} \in \mathbb{R}^{257\times1024}}$ with a patch size of 14. 
To balance DINOv2's pre-trained knowledge from large-scale datasets with domain-specific adaptation for hand-object interactions, we freeze the earlier layers in DINOv2 and only fine-tune its last twelve transformer layers. 
We leverage both the hand feature ${\rm f_{h}}$ and the image feature ${\rm f_{v}}$ for the following hand-held object reconstruction.

\subsection{Sparse point cloud reconstruction}
\label{sec:sparse_recon}

We reconstruct the hand-held object under the hand palm coordinate system, as the hand palm can be reliably localized by hand pose estimation models~\cite{pavlakos2024reconstructing,potamias2024wilor}.
To achieve this, we decompose the reconstruction task into two sub-tasks: canonical object point cloud generation and hand-relative object pose estimation. This disentanglement separates object shape recovery from pose estimation, facilitating the learning process. 
For object pose estimation, we predict only the object's 3D translation relative to the hand palm, denoted as ${\rm t_{o}} \in \mathbb{R}^{3}$. This is because many everyday objects exhibit high degrees of symmetry, making rotation prediction an ill-posed problem, as shown in previous works~\cite{chen2023gsdf,fu2024d}.
For sparse point cloud generation, we then produce the sparse object point cloud ${\rm p_{o}^{s}} \in \mathbb{R}^{\rm N^s_p\times3}$ in its rotated state, where ${\rm N^s_p}$ denotes the number of sparse points.

The sparse point cloud decoder employs a unified transformer backbone to jointly predict the object pose ${\rm t_{o}}$ and the rotated point cloud ${\rm p_{o}^{s}}$. 
Specifically, we define ${\rm 1+N^s_p}$ learnable token embeddings: one token for translation prediction and ${\rm N^s_p}$ tokens for point cloud generation.
These tokens are processed by a multi-layer transformer model, which first applies self-attention among the tokens, followed by cross-attention to condition on the extracted image features ${\rm f_{v}}$ and hand features ${\rm f_{h}}$.
The attention mechanism allows effective feature fusion across different modalities.
The output embedding of the first token is used to predict hand-relative object translation ${\rm t_{o}}$, while the remaining ${\rm N^s_p}$ output embeddings generate point cloud coordinates ${\rm p_{o}^{s}}$.
By sharing a single transformer backbone for predicting object pose and point cloud, the decoder effectively leverages mutual dependencies between the two tasks and is more efficient compared to prior work~\cite{fu2024d} that trains a separate object pose estimation network.

\begin{figure}[t]
  \centering
\includegraphics[width=0.95\linewidth]{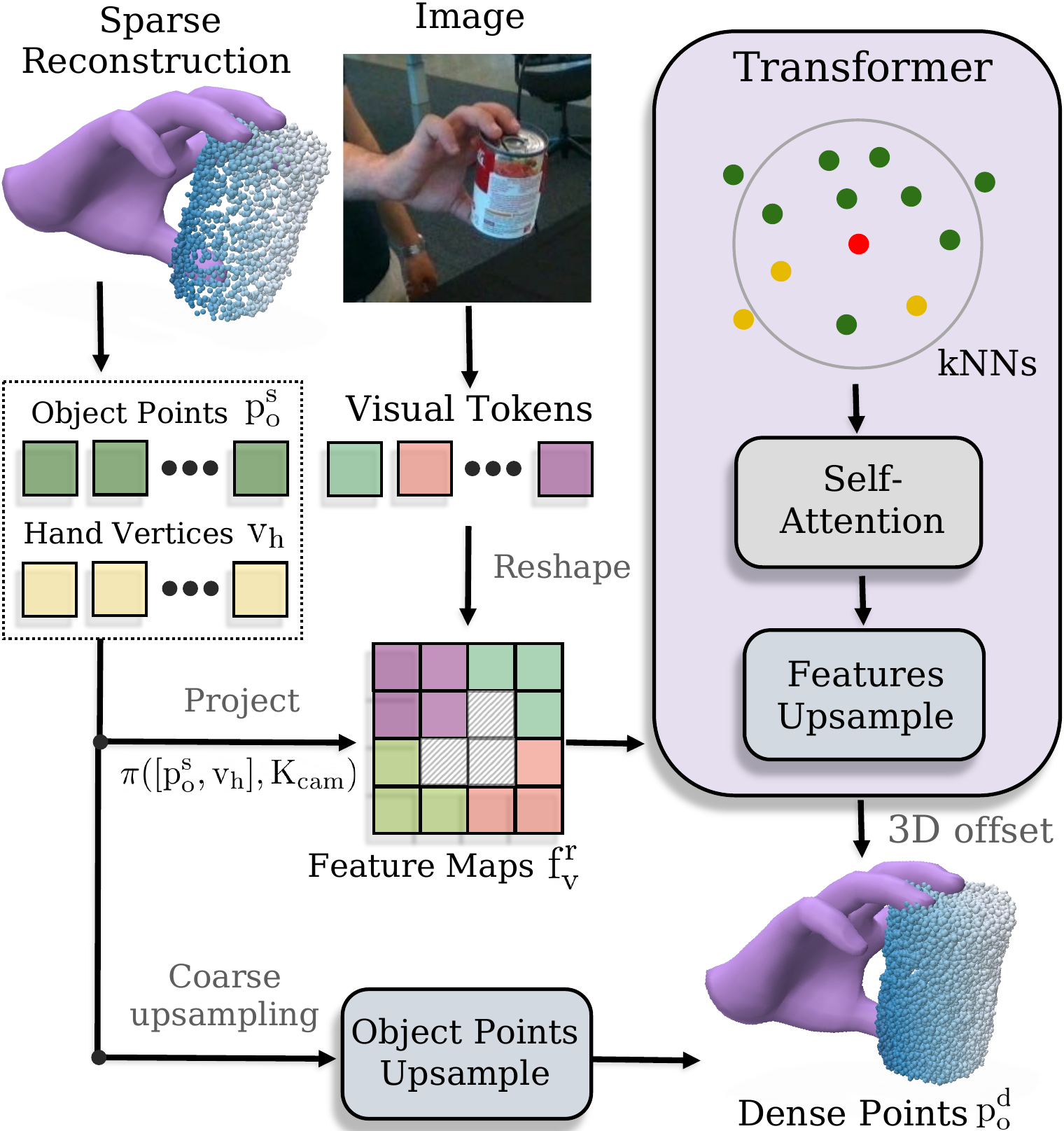}
\vspace{-0.3cm}
  \caption{
  Illustration of our dense point cloud decoder. The green and yellow points in the transformer indicate object points and hand vertices respectively. The model retrieves pixel-aligned image features for each reconstructed point and enhances its local context through self-attention. It upsamples the sparse point cloud to a high-resolution 3D object point cloud. 
  }
  \label{fig:upsample}
  \vspace{-0.5cm}
\end{figure}

\subsection{Dense point cloud reconstruction}
\label{sec:dense_recon}

Given the predicted sparse point clouds ${\rm p_{o}^{s}}$ and the hand-relative object pose ${\rm t_{o}}$, the dense point cloud decoder  further upscales it to produce high-resolution point cloud of the hand-held object. 
Figure~\ref{fig:upsample} provides details of the proposed dense point cloud decoder.

We first bilinearly upsample the reconstructed point clouds $\rm p_{o}^{s}$. However, the bilinearly upsampled points only provide a coarse approximation of the dense geometry, lacking alignment with image features and local contextual neighborhood information for each point. Therefore, we introduce an additional pixel-aligned feature extraction method that provides localized visual information to the dense point clouds decoder.
Specifically, we first exclude the global $\rm [CLS]$ token from $\rm f_{v} \in \mathbb{R}^{257\times1024}$ to focus only on the patch-level tokens. These remaining tokens are then reshaped into a spatial grid format and processed through $3\times3$ convolution layers, yielding a refined feature map $\rm {f^r_v}\in \mathbb{R}^{16\times16\times128}$. 
Next, we transform ${\rm p_{o}^{s}}$ from the object canonical space to the camera coordinate space using the estimated hand palm location ${\rm t_{p}}$ and hand-relative object translation ${\rm t_{o}}$. 
Using the predicted camera parameters ${\rm K_{cam}}$, we project ${\rm p_{o}^{s}}$ onto the image plane and retrieve pixel-aligned features for each point via perspective projection $\pi(\cdot)$ and bilinear interpolation $\rm F(\cdot)$:
\begin{equation}
{\rm f_{o}}= {\rm F}(\pi({\rm p_{o}^{s}} + {\rm t_{p}} + {\rm t_{o}},~{\rm K_{cam}}),~{\rm f_{v}^{r}}).
\label{eq:proj}
\end{equation}

The coordinates in the sparse point cloud ${\rm p_{o}^{s}} \in \mathbb{R}^{\rm N^s_p \times 3}$ are concatenated with their aligned visual features ${\rm f_{o}} \in \mathbb{R}^{\rm N^s_p \times 128}$. Similarly, we also retrieve visual features for 3D hand vertices ${\rm v_{h}}$ and feed them into the transformer model with object features. The model consists of two blocks, each performing self-attention followed by feature upsampling. The point cloud resolution is increased by factors of 2 and 4 in the first and second blocks, respectively.
Specifically, as shown in Figure~\ref{fig:upsample}, each block first applies self-attention~\cite{vaswani2017attention} within local neighborhoods to aggregate spatial and visual hand-object contexts for each object point, inspired by~\cite{xiang2021snowflakenet}. The neighborhoods are determined using $\rm k$ Nearest Neighbors (kNNs) based on spatial locations of the points, with $\rm k=16$ set empirically.
The point features are then bilinearly upsampled and passed through three convolutional layers to predict 3D offsets for each upsampled point.  
This process ultimately reconstructs a high-resolution dense point cloud ${\rm p_{o}^{d}} \in \mathbb{R}^{\rm N^d_p \times 3}$.

\subsection{Training}
\label{sec:train}

Unlike D-SCO~\cite{fu2024d} that trains each component separately, we adopt an end-to-end training strategy to optimize the entire HORT model. 
All components are jointly trained using an overall loss function $\mathcal{L}$ combining object pose loss along with sparse and dense point cloud generation losses:
\begin{equation}
\begin{split}
\mathcal{L} = \lambda_{1}\times\mathcal{L}_{pose} + \lambda_{2}\times\mathcal{L}_{cd}^{s} + \mathcal{L}_{cd}^{d}.
\end{split}
\label{loss_shape}
\end{equation}
where $\lambda_1$ and $\lambda_2$ are hyper-parameters to balance losses.

Specifically, we use the $\ell 1$ loss $\mathcal{L}_{pose}$ to predict the 3D object translation ${\rm t_o}$ relative to the hand palm as follows:
\begin{equation}
\mathcal{L}_{pose} = \Big \| {\rm t_o} - {\rm \hat{t}_o} \Big \|^1_1,
\label{eq:pose_loss}
\end{equation}
where ${\rm \hat{t}_o}$ denotes the ground-truth 3D offset between the object centroid and the palm. 

We employ the Chamfer Distance loss $\mathcal{L}_{cd}^{s}$ to supervise 3D point clouds generation as follows:
\begin{equation}
\mathcal{L}_{cd}^{\{s, d\}} = {\rm Chamfer~Distance}({\rm p_{o}^{\{s, d\}}},~{\rm \hat{p}_{o}^{\{s, d\}}}),
\label{eq:spc_loss}
\end{equation}
where ${\rm p_{o}^{\{\cdot\}}}$ denote either sparse (${\rm p_{o}^{s}}$) or dense (${\rm p_{o}^{d}}$) object point clouds. ${\rm \hat{p}_{o}^{\{\cdot\}}}$ represent their counterparts evenly sampled from the ground-truth object mesh.

%% file: sec/4_expr.tex
\section{Experiments}
\label{sec:experiment}
We conduct comprehensive experiments to validate the effectiveness of our approach on four diverse hand-held object reconstruction benchmarks. 

\subsection{Datasets}
\label{subsec:data}
\noindent \textbf{ObMan}~\cite{hasson2019learning}: It is a large-scale synthetic hand-object interaction dataset, containing 2,772 object instances from ShapeNet~\cite{chang2015shapenet} with 21K hand grasps generated by GraspIt!~\cite{miller2004graspit}. For fair comparison with previous works, we use the standard split which includes 141K and 6,285 samples for training and testing respectively. Following~\cite{chen2022alignsdf}, we randomly sample 30K samples from the standard training set for ablation experiments, while the full training data is used for comparison with state-of-the-art methods.

\noindent \textbf{HO3D}~\cite{hampali2020honnotate}: We use the HO3Dv3 version~\cite{hampali2021ho} of this dataset, which captures 68 videos sequences of ten subjects manipulating ten YCB objects~\cite{calli2015ycb}. 
We follow the standard experimental setup in \cite{ye2022s,zhang2024ddf,fu2024d} and include 77K samples for training and 1,221 samples for evaluation.

\noindent \textbf{DexYCB}~\cite{chao2021dexycb}: It captures hand grasping of twenty YCB objects from eight cameras by twenty subjects. Following the convention in ~\cite{yang2022artiboost,chen2022alignsdf,chen2023gsdf}, we focus on the right-hand samples and use the official s0 split. We follow the gSDF~\cite{chen2023gsdf} setup to downsample the video every 5 frames, resulting in 29K training images and 5,928 testing samples.

\noindent \textbf{MOW}~\cite{cao2021reconstructing}: It contains 520 images sampled from YouTube videos in 100 Days of Hands dataset~\cite{shan2020understanding}. The training and testing split contain 350 and 92 images respectively.

\subsection{Evaluation metrics}
\label{subsec:metric}
Following prior works~\cite{hasson2019learning,ye2022s,chen2023gsdf}, we use two main metrics to evaluate the 3D reconstruction quality. We also report interaction metrics to provide more details about the plausibility of our reconstructed hand-object configurations.

\noindent \textbf{Chamfer Distance (${\rm {\bf{CD}}}$)}. We compute Chamfer Distance (${\rm {cm}^2}$) between the reconstructed dense object point clouds and the ground-truth object point clouds. Following the standard protocol from previous works~\cite{chen2023gsdf}, we sample 30K points on both point clouds to compute Chamfer distance. We report the mean Chamfer distance over the whole test set to evaluate the quality of our reconstructed point clouds.

\noindent \textbf{F-score (${\rm {\bf{FS}}}$)}. Since Chamfer distance is sensitive to outliers~\cite{tatarchenko2019single,ye2022s}, we also report the F-score, a geometric mean of precision and recall, to comprehensively evaluate the reconstructed object point clouds and report F-score at 5mm (${\rm {FS}@5}$) and 10mm (${\rm {FS}@10}$) thresholds.

\noindent \textbf{Contact Ratio (${\rm \bf{Cr}}$)}. We follow previous works~\cite{karunratanakul2020grasping,chen2022alignsdf,chen2023gsdf} to report the percentage of testing samples where contact points between the reconstructed hand and object exist.

\noindent \textbf{Penetration Depth (${\rm \bf{Pd}}$)}. Following~\cite{karunratanakul2020grasping,chen2022alignsdf,chen2023gsdf}, we compute the maximum distance (in cm) from the estimated hand mesh vertices to the object's surface in case of a collision.

\input{table/ablation_encoder}
\input{table/ablation_decoder}

\subsection{Implementation details}
\label{subsec:details}
\noindent \textbf{Model architecture.} 
We use the large DINOv2 model~\cite{oquab2023dinov2} as our image encoder,  which consists of 24 transformer layers. 
The PointNet~\cite{qi2017pointnet} for hand feature extraction contains 5 MLP layers. 
The sparse point cloud decoder consists of 10 transformer layers with 8 attention heads. The learnable token for each point has 512 dimensional features. The initially decoded sparse object point clouds contains $\rm N^s_p=2,048$ points. Followed by 3 layers of point deconvolution and transformer layers, we finally reconstruct a dense point cloud of $\rm N^d_p=16,384$ object points, which is the same number of points used in D-SCO~\cite{fu2024d}.

\noindent \textbf{Training details.} 
We crop the hand-object region out of the original image like previous works~\cite{chen2022alignsdf,hasson2019learning,ye2022s}. In training, according to~\cite{mehta2017monocular,yu2021pcls}, we adjust camera intrinsics and extrinsics according to the bounding box location and resizing ratio.
The input image size for our model is $224\times224$. We perform data augmentation including random rotation ($[-90^{\circ}, 90^{\circ}]$), random resizing ($[0.8, 1.2]$), random bounding box shifting and color jittering. During training, we use ground-truth camera parameters and hand poses, but our approach does not assume that they are available at test time. 
We train our models with a batch size of 192 for 50 epochs on both ObMan and DexYCB datasets. We use the Adam optimizer~\cite{kingma2014adam} with a base learning rate of 1e-4 and the cosine learning rate decay. We empirically set both $\lambda_{1}$ and $\lambda_{2}$ to two. We fine-tune our ObMan pre-trained model on HO3D, DexYCB and MOW datasets for a fair comparison with previous works. The training takes around 50 hours on ObMan dataset with 4 NVIDIA A100 GPUs.

\subsection{Ablation studies}
\label{subsec:ablation}
We conduct extensive ablation studies on the ObMan dataset to evaluate the effectiveness of each component in our model. 
We follow~\cite{chen2022alignsdf} to use randomly sampled 30K data for training and the original test set for evaluation. 

\input{table/sota_obman_new}
\input{table/sota_ho3d_dexycb}

\begin{figure*}[t]
  \centering
\includegraphics[width=0.95\linewidth]{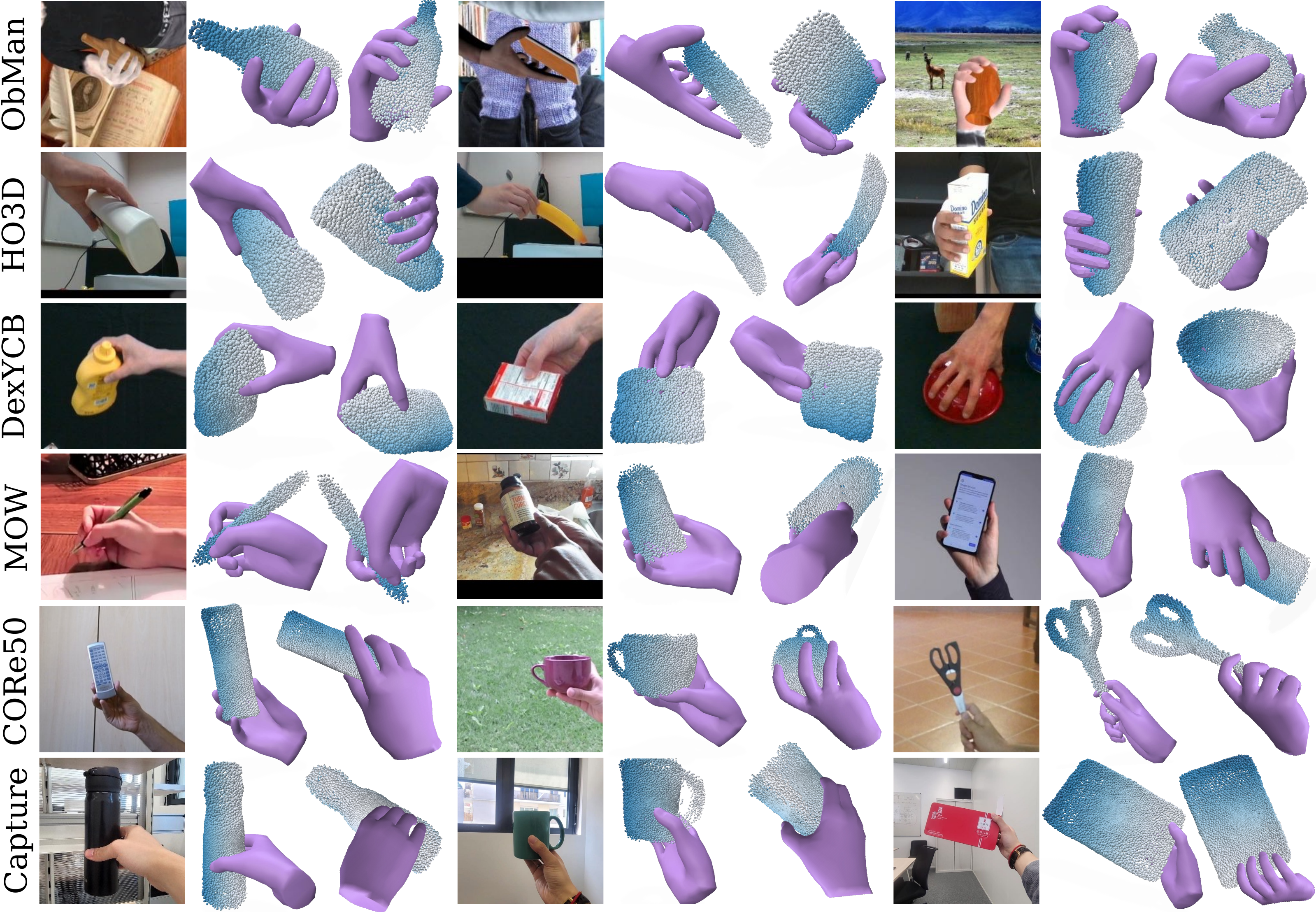}
\vspace{-0.4cm}
\caption{Qualitative results of the proposed model on diverse benchmarks. The first row shows our results on the synthetic ObMan dataset. The second, third and forth rows present our results on HO3D, DexYCB and MOW datasets respectively. The fifth row illustrates our performance on in-the-wild CORe50 images~\cite{lomonaco2017core50}. The last row shows the results of our model on images captured by our mobile phone. Our approach can accurately reconstruct point clouds of hand-held objects from both synthetic and real-world images.}
\vspace{-0.4cm}
\label{fig:demo}
\end{figure*}

\noindent \textbf{Hand encoder.} 
The first three rows in Table~\ref{tab:ablation_encoder} evaluate the contributions of our proposed hand encoder.
The base model in R1, which relies solely on image features without a hand encoder, achieves poor results.
R2 incorporates the 3D hand vertices under the hand palm coordinate system as input, which significantly improves performance across all metrics, for example, the ${\rm {FS}@5}$ improves from 0.45 to 0.53 and ${\rm CD}$ from 3.1 to 2.4. In addition, it also greatly improves the contact ratio $\rm Cr$ and reduces the penetration depth $\rm Pd$. This confirms that hand information substantially enhances reconstruction accuracy of hand-held objects and better recover interaction between the hand and the object.  Further enhancement of hand features in R3, where hand vertices are transformed into multiple hand joint coordinate systems, yields additional gains particularly for ${\rm {FS}@5}$ increased from 0.53 to 0.60.
These results clearly demonstrate the necessity of effectively encoding hand geometry for accurate hand-held object reconstruction.

\noindent \textbf{Image encoder.} 
In the last three rows of Table~\ref{tab:ablation_encoder}, we compare different training strategies for the image encoder.
R4 trains a ViT image encoder~\cite{wu2024reconstructing} from scratch on the ObMan dataset, while R5 uses the frozen DINOv2 model. Both approaches perform worse than our fine-tuning strategy in R3 which refines the visual features for the hand-object interaction domain.

\noindent \textbf{Sparse point cloud decoder.} 
We evaluate our joint decoding strategy for the sparse point cloud decoder in the first two rows of Table~\ref{tab:ablation_decoder}.
The model in R1 employs two separate modules: one for object pose prediction, which concatenates the global image feature (\emph{i.e.}, $\rm [CLS]$ token in DINOv2) with 3D hand features from PointNet as input to an MLP for translation prediction; and another for point cloud generation similar to our final sparse decoder.
In contrast, the model in R2 jointly predicts the object pose and shape through a unified transformer decoder. 
This design integrates object volume features for more accurate object localization, leading to enhanced performance across metrics.

\noindent \textbf{Dense point cloud decoder.} 
We further evaluate our coarse-to-fine strategy and dense point cloud decoder design in the last four rows of Table~\ref{tab:ablation_decoder}.
In R3, we concatenate coordinates of the sparse point cloud with global image features before feeding them to the dense decoder. Compared to R2, which lacks the point cloud upsampling step, R3 shows significant improvement in reconstruction performance. R4 further replaces the global image features used in R3 with pixel-aligned features and improves the performance. As shown in Figure~\ref{fig:upsample}, R5 additionally includes hand features and integrate both hand and object contexts in the transformer model for object points upsampling, which achieves the best results across all evaluation metrics.

\subsection{Comparison with state of the art}
\label{subsec:sota}

We compare our approach with state-of the-art methods on four benchmarks.
Figure~\ref{fig:demo} presents some qualitative examples of our HORT model on these datasets.

\noindent \textbf{ObMan.}
Table~\ref{tab:sota_obman} presents a comprehensive comparison on ObMan dataset, where we evaluate models under two testing scenarios: using either ground-truth or estimated hand poses as input. 
For the estimated hand poses, we follow the same practice as in previous works~\cite{ye2022s,fu2024d} to train a hand pose estimation model~\cite{hasson2019learning} for ObMan. Our approach consistently outperforms state-of-the-art methods under both settings. 
Our model shows substantial improvement over methods based on implicit functions (\emph{i.e.}, DDF-HO~\cite{zhang2024ddf} and gSDF~\cite{chen2023gsdf}) and the recent D-SCO~\cite{fu2024d} method.

\noindent \textbf{HO3D.} In the left block of Table~\ref{tab:sota_ho3d_dexycb}, we compare different methods on the real-world HO3D benchmark. 
We use predicted hand poses and camera parameters from an off-the-shelf hand model~\cite{potamias2024wilor} and fine-tune the model from the ObMan pre-trained model for fair comparison with prior works~\cite{zhang2024ddf,ye2022s,fu2024d}. Our model constantly outperforms previous methods under all evaluation metrics. Compared with D-SCO~\cite{fu2024d}, our approach achieves relative improvement of 14.3\% and 23.5\% for ${\rm {FS}@10}$ and ${\rm {{CD}}}$ respectively.

\noindent \textbf{DexYCB.} The right block of Table~\ref{tab:sota_ho3d_dexycb} presents our experimental results on the DexYCB dataset. We follow prior works~\cite{zhang2024ddf,fu2024d} to use a hand model~\cite{qi2024hoisdf} that has been specifically trained on this dataset. As shown in Table~\ref{tab:sota_ho3d_dexycb}, our model outperforms the strong video-based gSDF model~\cite{chen2023gsdf} and the state-of-the-art D-SCO method~\cite{fu2024d},  achieving better ${\rm {FS}@10}$ and ${\rm {{CD}}}$.

\input{table/sota_mow}

\begin{figure}[!t]
\centering
\includegraphics[width=0.9\linewidth]{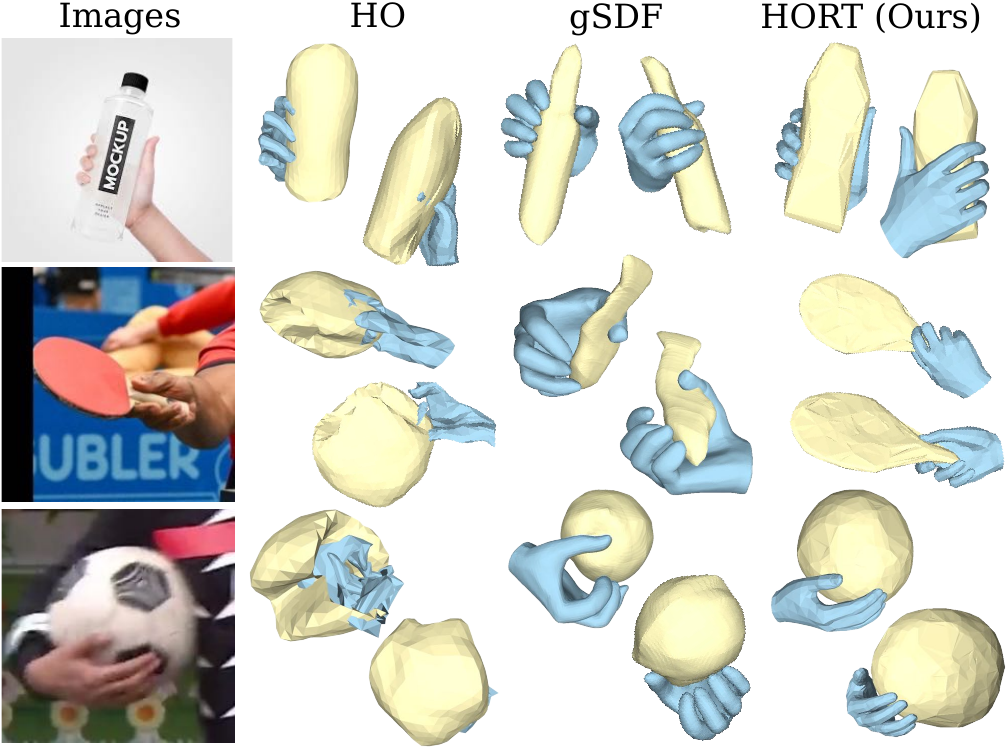}
\vspace{-0.3cm}
\caption{Qualitative comparison with previous methods~\cite{hasson2019learning,chen2023gsdf}.}
\label{sota:qualitative_comparison}
\vspace{-0.7cm}
\end{figure}

\noindent \textbf{MOW.} We conduct experiments on the MOW dataset under two settings for fair comparison with previous works. In the first setting~\cite{fu2024d,ye2022s,zhang2024ddf}, we pre-train the model on ObMan and fine-tune it on MOW. In the second setting~\cite{wu2024reconstructing}, we train our model jointly on the DexYCB, HOI4D~\cite{liu2022hoi4d} and MOW datasets. As shown in Table~\ref{tab:sota_mow}, our model outperforms previous methods in both settings, demonstrating the robustness of our method on challenging YouTube samples.

\noindent \textbf{Generalization to in-the-wild images.} Figure~\ref{fig:demo} shows the performance of our model on CORe50 images~\cite{lomonaco2017core50} ($5_{th}$ row) and images captured by our phone ($6_{th}$ row). It demonstrates that our model generalizes robustly to diverse objects and textures after joint training on the DexYCB, HOI4D, and MOW datasets, as shown in Table~\ref{tab:sota_mow}. We also qualitatively compare HORT with previous approaches~\cite{hasson2019learning,chen2023gsdf} in Figure~\ref{sota:qualitative_comparison}, which demonstrates that HORT produces more convincing results. Please see appendix for our robustness to occlusion and failure cases analysis.

\input{table/sota_runtime_v2}

\noindent \textbf{Runtime comparison.}
As shown in Table~\ref{tab:sota_runtime}, we compare the runtime and computational cost of our method with previous approaches on the ObMan dataset. Inference speed and GFLOPs are measured on an NVIDIA A100 GPU for hand-held object reconstruction including hand prediction. Implicit methods require inference on densely sampled points (\emph{i.e.}, $128\times128\times128$ in common practice) at test time, resulting in over 6000 GFLOPs. D-SCO~\cite{fu2024d} is even slower than implicit models due to intensive denoising steps. 
HO~\cite{hasson2019learning} is very fast but suffers from low-quality 3D reconstruction.
Our model requires only 258 GFLOPs and achieves robust performance, running at 16.9 fps including both 3D hand and object reconstruction.

\noindent \textbf{Point clouds versus meshes.} Given the reconstructed point clouds, fitting meshes using off-the-shelf algorithms~\cite{edelsbrunner1983shape,kazhdan2006poisson} is straightforward. This process does not affect performance; \eg, {\rm {FS}@10} remains at 0.85 on the DexYCB dataset.
However, it increases the runtime, reducing the speed from 16.9 fps to 4.8 fps. Despite this, our model remains significantly faster than methods based on implicit functions and the recent D-SCO~\cite{fu2024d} method in Table~\ref{tab:sota_runtime}.
\vspace{-0.2cm}

%% file: table/ablation_encoder.tex
\begin{table}[t]
\centering
\footnotesize
\caption{The performance of our hand-held object reconstruction models with variants of visual feature encoders on ObMan dataset.}
\vspace{-0.3cm}
\setlength{\tabcolsep}{2pt}
\renewcommand\arraystretch{0.9}
\begin{tabular}{ccccccccc}
\toprule
\multirow{2}{*}{} &\multicolumn{2}{c}{Hand Encoder} &\multirow{2}{*}{\begin{tabular}[c]{@{}c@{}}Image\\ Encoder\end{tabular}}& \multirow{2}{*}{${\rm {FS}@5}\uparrow$} & \multirow{2}{*}{${\rm {FS}@10}\uparrow$}  & \multirow{2}{*}{${\rm {CD}}\downarrow$} & \multirow{2}{*}{${\rm {Cr}}\uparrow$}&\multirow{2}{*}{${\rm {Pd}}\downarrow$}\\ \cmidrule(lr){2-3}
 &Palm&Joints& &  &  & & &  \\ \midrule
R1&$\times$&$\times$&Fine-tune&0.45&0.75&3.1&0.76&2.15 \\
R2&\checkmark&$\times$&Fine-tune&0.53&0.82&2.4&0.85&1.76 \\
R3&\checkmark&\checkmark&Fine-tune&\textbf{0.60}&\textbf{0.87}&\textbf{1.8}&\textbf{0.92}&\textbf{1.37} \\
R4&\checkmark&\checkmark&Scratch&0.51&0.79&2.6&0.88&1.48\\
R5&\checkmark&\checkmark&Frozen&0.48&0.76&2.9&0.87&1.62\\ 
\bottomrule
\end{tabular}
\vspace{-0.3cm}
\label{tab:ablation_encoder}
\end{table}

%% file: table/ablation_decoder.tex
\begin{table}[t]
\centering
\footnotesize
\caption{The performance of our hand-held object reconstruction models with various object decoder variants on ObMan dataset. J-decode indicates models that jointly decode sparse point clouds with hand-relative 3D translation using a transformer architecture.
}
\vspace{-0.3cm}
\setlength{\tabcolsep}{1pt}
\renewcommand\arraystretch{0.9}
\begin{tabular}{c c ccc ccc}
\toprule
\multirow{2}{*}{} & \multicolumn{1}{c}{Sparse} & \multicolumn{3}{c}{Dense} &\multirow{2}{*}{${\rm {FS}@5}\uparrow$}&\multirow{2}{*}{${\rm {FS}@10}\uparrow$}&\multirow{2}{*}{${\rm {CD}}\downarrow$} \\
\cmidrule(lr){2-2} \cmidrule(lr){3-5}
 & J-decode &upsample&aligned feat.&hand feat.&&& \\ \midrule
R1&$\times$&$\times$&$\times$&$\times$&0.41&0.69&3.8\\
R2&\checkmark&$\times$&$\times$&$\times$&0.47&0.77&3.0\\
R3&\checkmark&\checkmark&$\times$&$\times$&0.54&0.82&2.3\\
R4&\checkmark&\checkmark&\checkmark&$\times$&0.58&0.85&2.0\\
R5&\checkmark&\checkmark&\checkmark&\checkmark&\textbf{0.60}&\textbf{0.87}&\textbf{1.8}\\
\bottomrule
\end{tabular}
\vspace{-0.6cm}
\label{tab:ablation_decoder}
\end{table}

%% file: table/sota_obman_new.tex
\begin{table}[!t]
\centering
\footnotesize
\caption{Comparison with state-of-the-art methods on ObMan dataset. HORT achieves superior performance across all metrics when using either predicted or ground-truth 3D hands.}
\vspace{-0.3cm}
\setlength{\tabcolsep}{2pt} 
\renewcommand\arraystretch{0.9} 
\begin{tabular}{l ccc ccc}
\toprule
\multirow{2}{*}{Methods} & \multicolumn{3}{c}{Predicted hand} & \multicolumn{3}{c}{Ground-truth hand} \\
\cmidrule(lr){2-4} \cmidrule(lr){5-7}
 & ${\rm {FS}@5}\uparrow$ & ${\rm {FS}@10}\uparrow$ &${\rm {CD}}\downarrow$& ${\rm {FS}@5}\uparrow$ & ${\rm {FS}@10}\uparrow$ &${\rm {CD}}\downarrow$\\ 
\midrule
HO~\cite{hasson2019learning}&0.23&0.56&6.4&-&-&-\\ 
GF~\cite{karunratanakul2020grasping}&0.30&0.51&13.9&-&-&-\\
IHOI~\cite{ye2022s}&0.42&0.63&10.2&0.49&0.70&9.2\\
AlignSDF~\cite{chen2022alignsdf}&0.40&0.64&9.2&-&-&-\\
gSDF~\cite{chen2023gsdf}&0.44&0.66&8.8&-&-&-\\
DDF-HO~\cite{zhang2024ddf}&0.55&0.67&1.4&-&-&-\\ 
D-SCO~\cite{fu2024d}&0.61&0.81&1.1&0.65&0.85&1.0\\ 
HORT (Ours)&\textbf{0.66}&\textbf{0.88}&\textbf{1.0}&\textbf{0.72}&\textbf{0.91}&\textbf{0.9}\\ 
\bottomrule
\end{tabular}
\vspace{-0.25cm}
\label{tab:sota_obman}
\end{table}

%% file: table/sota_ho3d_dexycb.tex
\begin{table}[t]
\centering
\footnotesize
\caption{Comparison with state-of-the-art methods on HO3D and DexYCB. HORT achieves the best performance on both datasets. }
\vspace{-0.3cm}
\setlength{\tabcolsep}{2pt} 
\renewcommand\arraystretch{0.9} 
\begin{tabular}{l ccc ccc}
\toprule
\multirow{2}{*}{Methods} & \multicolumn{3}{c}{HO3D} & \multicolumn{3}{c}{DexYCB} \\
\cmidrule(lr){2-4} \cmidrule(lr){5-7}
 & ${\rm {FS}@5}\uparrow$ & ${\rm {FS}@10}\uparrow$ &${\rm {CD}}\downarrow$& ${\rm {FS}@5}\uparrow$ & ${\rm {FS}@10}\uparrow$ &${\rm {CD}}\downarrow$\\ 
\midrule
HO~\cite{hasson2019learning}&0.11&0.22&41.9&0.38&0.64&4.2\\ 
GF~\cite{karunratanakul2020grasping}&0.12&0.24&49.6&0.39&0.66&4.5\\
IHOI~\cite{ye2022s}&0.28&0.50&15.3&-&-&-\\
AlignSDF~\cite{chen2022alignsdf}&-&-&-&0.41&0.68&3.9\\
gSDF~\cite{chen2023gsdf}&-&-&-&0.44&0.71&3.4\\
DDF-HO~\cite{zhang2024ddf}&0.27&0.40&8.6&-&-&-\\ 
D-SCO~\cite{fu2024d}&0.41&0.63&3.4&\textbf{0.63}&0.82&1.3\\ 
HORT (Ours)&\textbf{0.45}&\textbf{0.72}&\textbf{2.6}&\textbf{0.63}&\textbf{0.85}&\textbf{1.1}\\ 
\bottomrule
\end{tabular}
\vspace{-0.55cm}
\label{tab:sota_ho3d_dexycb}
\end{table}

%% file: table/sota_mow.tex
\begin{table}[!t]
\centering
\footnotesize
\caption{Comparison with state-of-the-art methods on MOW. HORT achieves the best performance under both evaluation setup.}
\vspace{-0.3cm}
\setlength{\tabcolsep}{2pt} 
\renewcommand\arraystretch{0.9} 
\begin{tabular}{l ccc ccc}
\toprule
\multirow{2}{*}{Methods} & \multicolumn{3}{c}{Fine-tune from ObMan} & \multicolumn{3}{c}{Multiple datasets} \\
\cmidrule(lr){2-4} \cmidrule(lr){5-7}
 & ${\rm {FS}@5}\uparrow$ & ${\rm {FS}@10}\uparrow$ &${\rm {CD}}\downarrow$ & ${\rm {FS}@5}\uparrow$ & ${\rm {FS}@10}\uparrow$&${\rm {CD}}\downarrow$ \\ 
\midrule
IHOI~\cite{ye2022s}&0.13&0.24&49.8& - & -&- \\ 
DDF-HO~\cite{zhang2024ddf}&0.17&0.24&15.9& - & -&- \\ 
D-SCO~\cite{fu2024d}&\textbf{0.30}&0.47& 13.5 & -&-&- \\ 
WildHOI~\cite{prakash2023learning}& - & - & -&0.08 & 0.20&27.2 \\ 
MCC-HO~\cite{wu2024reconstructing}& - & - & -&0.15 & 0.31&15.2 \\ 
HORT (Ours)& \textbf{0.30} & \textbf{0.49} &\textbf{10.2} &\textbf{0.37} & \textbf{0.57}&\textbf{5.4} \\ 
\bottomrule
\end{tabular}
\label{tab:sota_mow}
\vspace{-0.35cm}
\end{table}

%% file: table/sota_runtime_v2.tex
\begin{table}[t]
\centering
\footnotesize
\caption{Runtime comparison with previous methods on ObMan. HORT significantly reduces computational cost and accelerates model inference. We report the inference speed of our HORT algorithm both with and without the meshing step.}
\setlength{\tabcolsep}{1pt}
\renewcommand\arraystretch{0.9}
\vspace{-0.3cm}
\begin{tabular}{lccccccc}
\toprule
Methods &\#points&\#verts&GFLOPs$\downarrow$&Params$\downarrow$&fps$\uparrow$&fps mesh$\uparrow$&${\rm CD}\downarrow$ \\ \midrule
HO~\cite{hasson2019learning} &N/A&0.6k&\textbf{76}&25M&\textbf{55.2}&\textbf{55.2}&6.4 \\
GF~\cite{karunratanakul2020grasping}&N/A&20k&6.4k&26M&0.37&0.37&13.9 \\
IHOI~\cite{ye2022s}&N/A&20k&5.9k&\textbf{24M}&0.37&0.37&10.2\\
AlignSDF~\cite{chen2022alignsdf}&N/A&20k&6.6k&33M&0.48&0.48&9.2\\
gSDF~\cite{chen2023gsdf}&N/A&20k&6.7k&35M&0.44&0.44&8.8\\
DDF-HO~\cite{zhang2024ddf}&N/A&20k&8.9k&25M&0.35&0.35&1.4\\
D-SCO~\cite{fu2024d}&16k&1k&-&-&0.08&-&1.1\\
HORT (Ours) &16k&1k&0.25k&366M&16.9&4.8&\textbf{1.0}\\ 
\bottomrule
\end{tabular}
\vspace{-0.5cm}
\label{tab:sota_runtime}
\end{table}

%% file: sec/5_conclusion.tex
\section{Conclusion}
\vspace{-0.1cm}
\label{sec:conslusion}

This work presents HORT, a transformer-based framework for reconstructing dense point clouds of hand-held objects from monocular RGB images. 
Our approach follows an efficient coarse-to-fine learning paradigm to first generate a sparse object point cloud and progressively refine it to a high-fidelity dense point cloud.
To further improve reconstruction accuracy, we propose to extract fine-grained 3D hand geometry features together with image information for joint object pose and point clouds prediction, as well as pixel-aligned features for upsampling.
Comprehensive experiments across four  benchmarks including ObMan, HO3D, DexYCB and MOW consistently demonstrate the superiority of our method over existing approaches.

{
\scriptsize
\renewcommand{\baselinestretch}{0.5}
\noindent \textbf{Acknowledgement}:
This work was granted access to HPC resources of IDRIS under the allocation AD011013147 made by GENCI. 
It was funded in part by the French government under management of Agence Nationale de la Recherche as part of the “France 2030" program, reference ANR-23-IACL-0008 (PR[AI]RIE-PSAI projet), and the ANR project VideoPredict (ANR-21-FAI1-0002-01) and the Paris Île-de-France Région in the frame of the DIM AI4IDF. R.A. Potamias was partially supported by EPSRC Project GNOMON (EP/X011364). Cordelia Schmid would like to acknowledge the support by the Körber European Science Prize.
}

%% file: appendix.tex
\appendix

In appendix, we provide more details of our model architectures and additional results. We first present details of our model architecture in Section~\ref{supmat:arch}. Then in Section~\ref{supmat:exp}, we discuss additional experimental results.

\section{Network Architecture}
\label{supmat:arch}
\subsection{PointNet encoder}

\begin{figure}[!ht]
  \centering
  \includegraphics[width=1.0\linewidth]{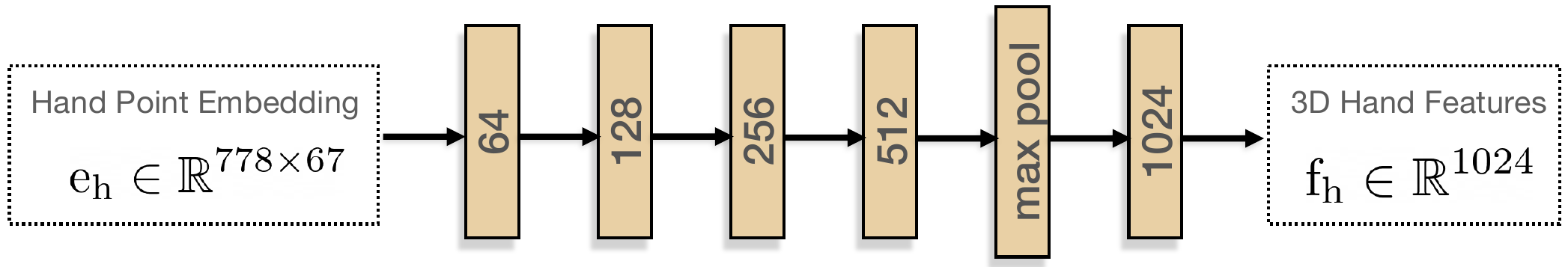}
  \vspace{-0.5cm}
  \caption{Network architecture used for our PointNet encoder. We use five fully-connected layers for efficiency. The number in the box denotes the dimension of features.}
  \label{fig:supmat_pointnet}
  \vspace{-0.3cm}
\end{figure}

As shown in Figure~\ref{fig:supmat_pointnet}, we employ five fully-connected layers to construct an efficient PointNet encoder. It takes the point clouds transformed into different hand coordinate systems (${\rm e_{h} \in \mathbb{R}^{778\times67}}$) and outputs 1024-dimensional global features ${\rm f_{h} \in \mathbb{R}^{1024}}$ for the sparse point cloud decoder.

\subsection{Dense point cloud decoder}
\begin{figure}[!ht]
  \centering
  \includegraphics[width=1.0\linewidth]{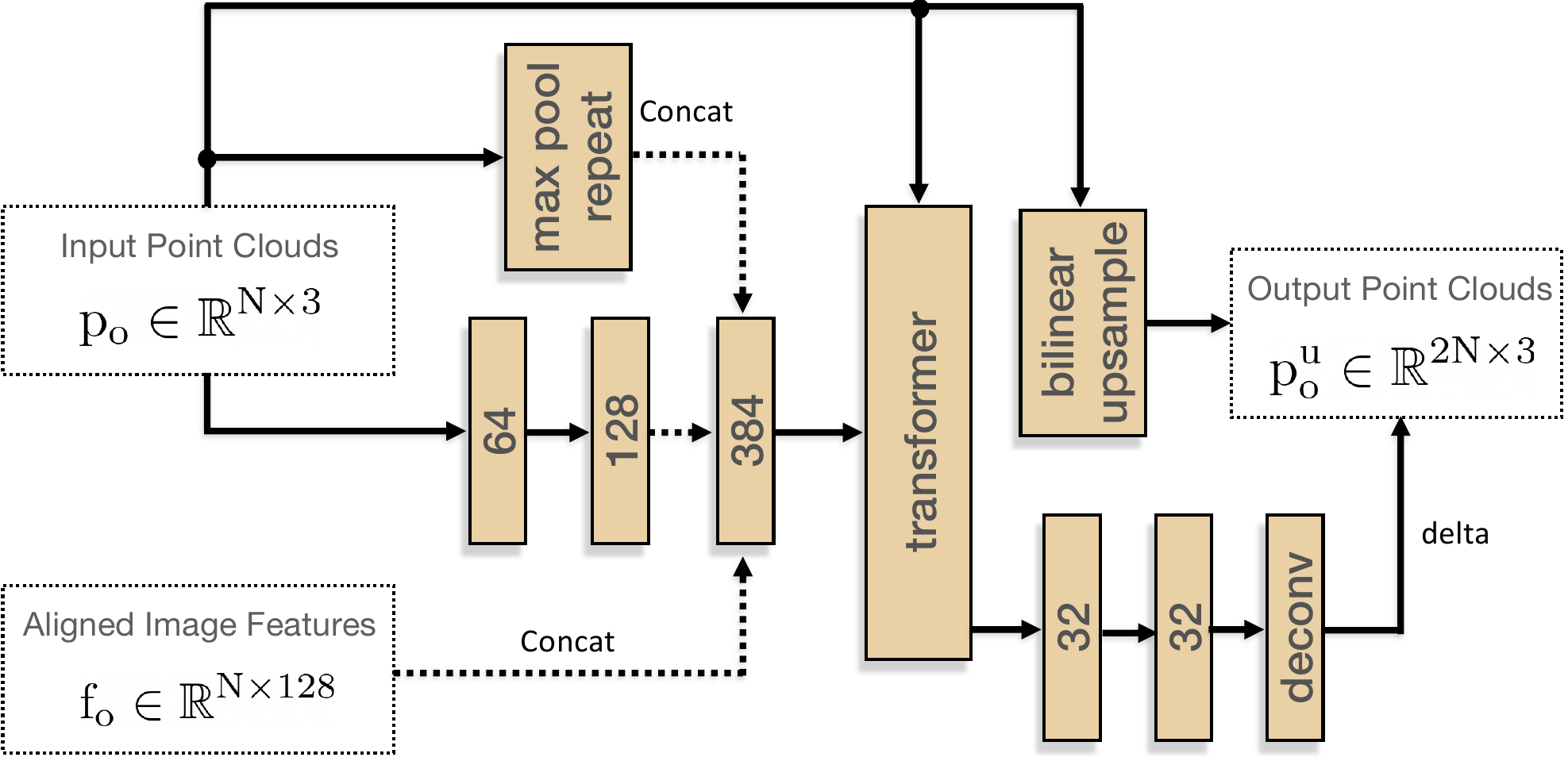}
  \vspace{-0.5cm}
  \caption{Network architecture used for reconstructing dense object point clouds. Given the input point clouds and its corresponding image features, the network performs point clouds upsampling. The number in the box denotes the dimension of features.}
  \label{fig:supmat_upsample}
  \vspace{-0.3cm}
\end{figure}

Figure~\ref{fig:supmat_upsample} illustrates the network architecture for reconstructing dense object point clouds. Given input point clouds $\rm {p_{o}} \in \mathbb{R}^{N\times3}$ and aligned image features $\rm {f_{o}} \in \mathbb{R}^{N\times128}$, we first apply convolution layers and max pooling to $\rm {p_{o}}$, then concatenate the processed features with $\rm {f_{o}}$. As discussed in the main paper, we perform the same operations for predicted hand vertices $\rm {v_{h}}$, equipping each object point with its surrounding hand context. Next, we upsample the input object point clouds via bilinear interpolation and employ a transformer model to predict 3D offsets for the initially upsampled object points. Our model consists of two consecutive blocks, as shown in Figure~\ref{fig:supmat_upsample}, which progressively upsample sparse point clouds $\rm {p_{o}^{s}} \in \mathbb{R}^{2048\times3}$ to dense object point clouds $\rm {p_{o}^{d}} \in \mathbb{R}^{16384\times3}$.

\begin{figure}[t]
  \centering
  \includegraphics[width=1.0\linewidth]{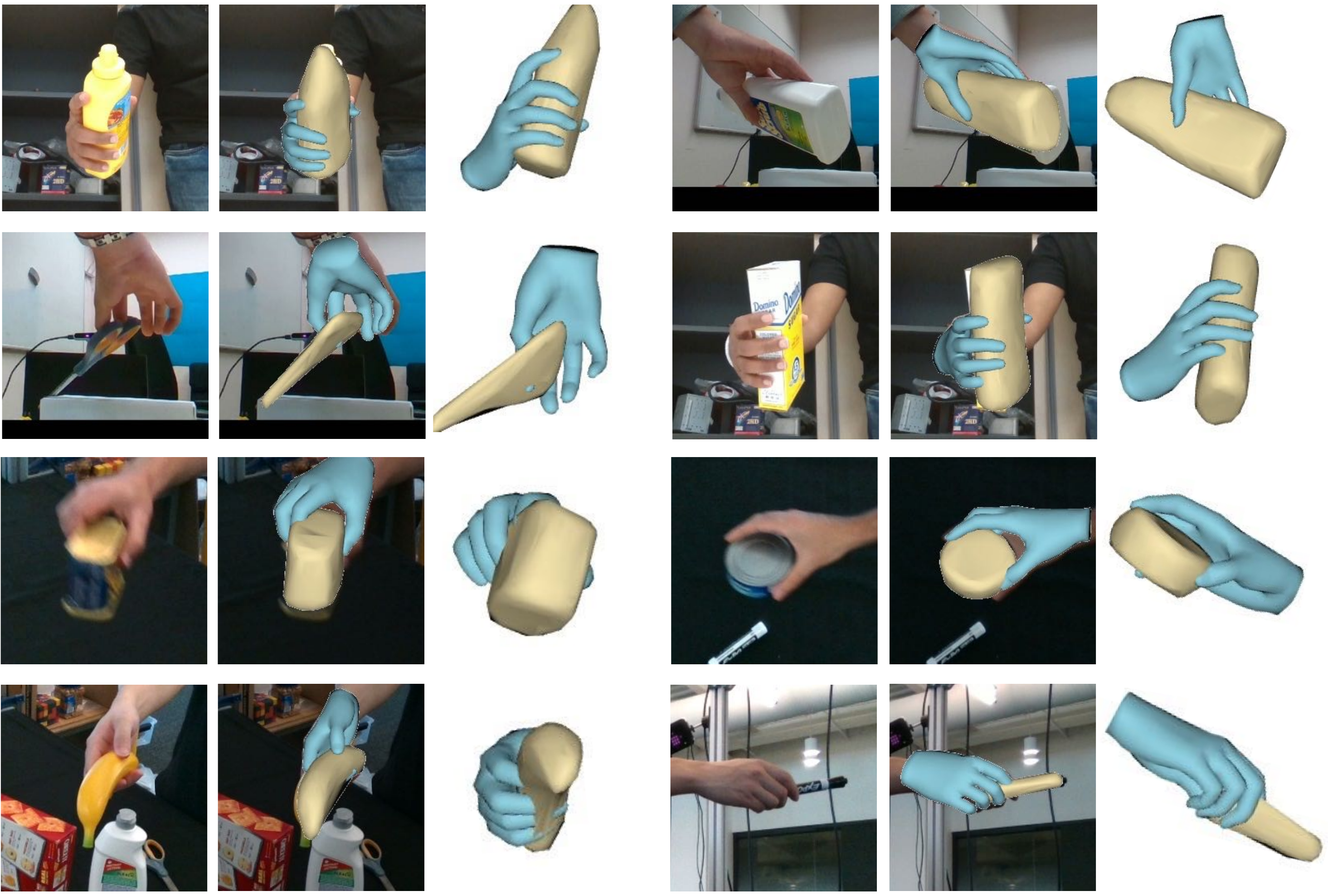}
  \vspace{-0.5cm}
  \caption{Qualitative results of reconstructed object meshes produced by our approach on HO3D~\cite{hampali2020honnotate} and DexYCB~\cite{chao2021dexycb} datasets.}
  \label{fig:supmat_mesh}
  \vspace{-0.4cm}
\end{figure}

\section{Experimental Results}
\label{supmat:exp}
\subsection{Object mesh reconstruction results}
When needed, we can easily convert our reconstructed object point clouds into meshes. As shown in Figure~\ref{fig:supmat_mesh}, we employ the alpha shapes algorithm~\cite{edelsbrunner1983shape}, as implemented in the Open3D library, to generate high-fidelity object meshes.~Table~\ref{tab:supmat_T1} presents our quantitative analysis of point cloud and mesh predictions. Here, we use predicted hand poses for different approaches on DexYCB~\cite{chao2021dexycb} dataset. Compared to evaluations on point clouds in R5,~R6 evaluates object meshes and achieves similar quantitative performance, further demonstrating the flexibility and efficiency of our point cloud representation. Moreover, compared to previous methods, HORT produces more physically plausible hand-object configurations and achieves better results in terms of Contact Ratio ($\rm Cr$) and Penetration Depth ($\rm Pd$).

\input{table/supmat_mesh}
\input{table/supmat_hand}
\vspace{-0.3cm}

\begin{figure}[t]
  \centering
 \includegraphics[width=1.0\linewidth]{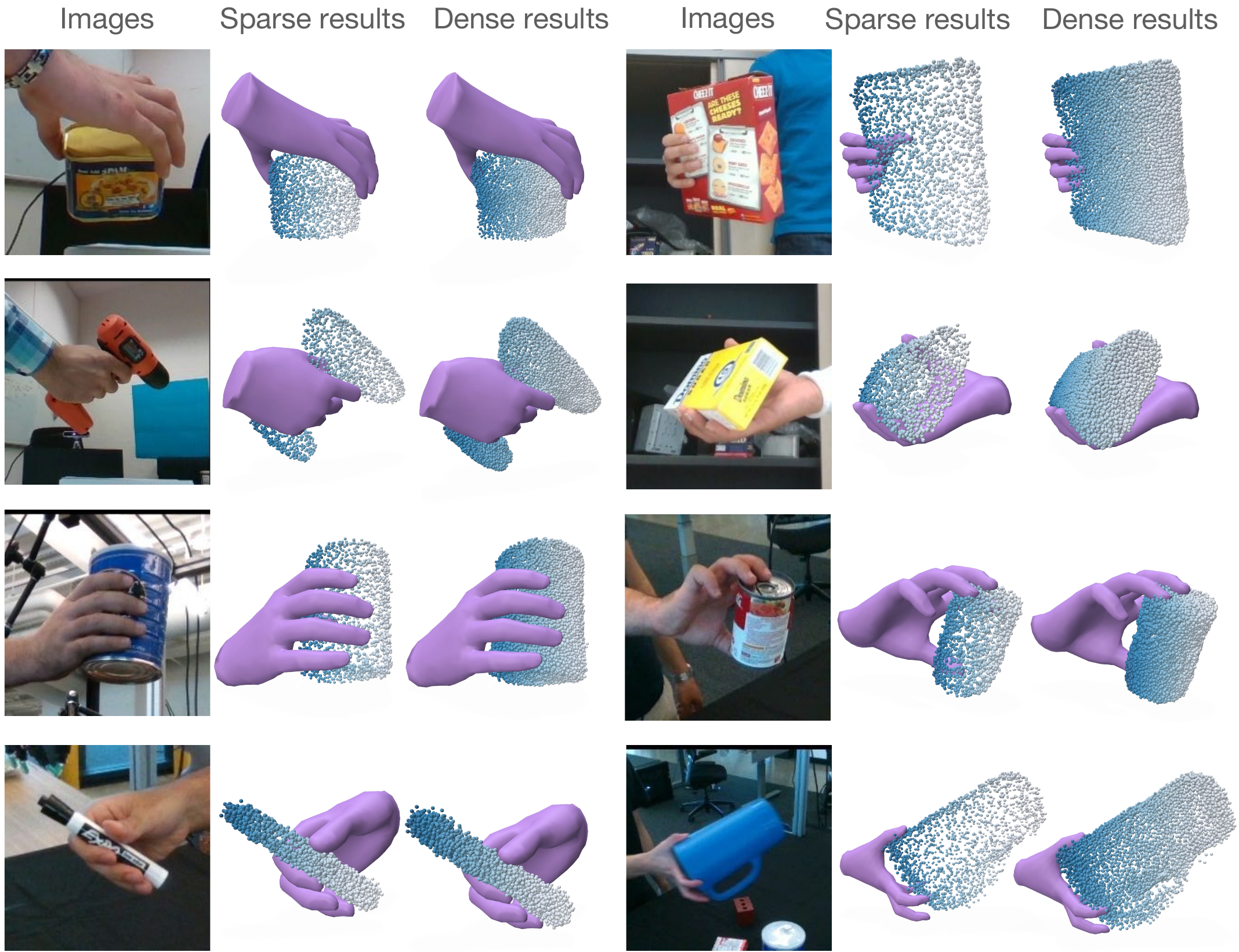}
 \vspace{-0.5cm}
  \caption{Qualitative comparison of our reconstructed sparse and dense object point clouds on HO3D and DexYCB datasets.}
  \label{fig:supmat_density}
  \vspace{-0.2cm}
\end{figure}

\begin{figure}[!t]
\centering
\includegraphics[width=1.0\linewidth]{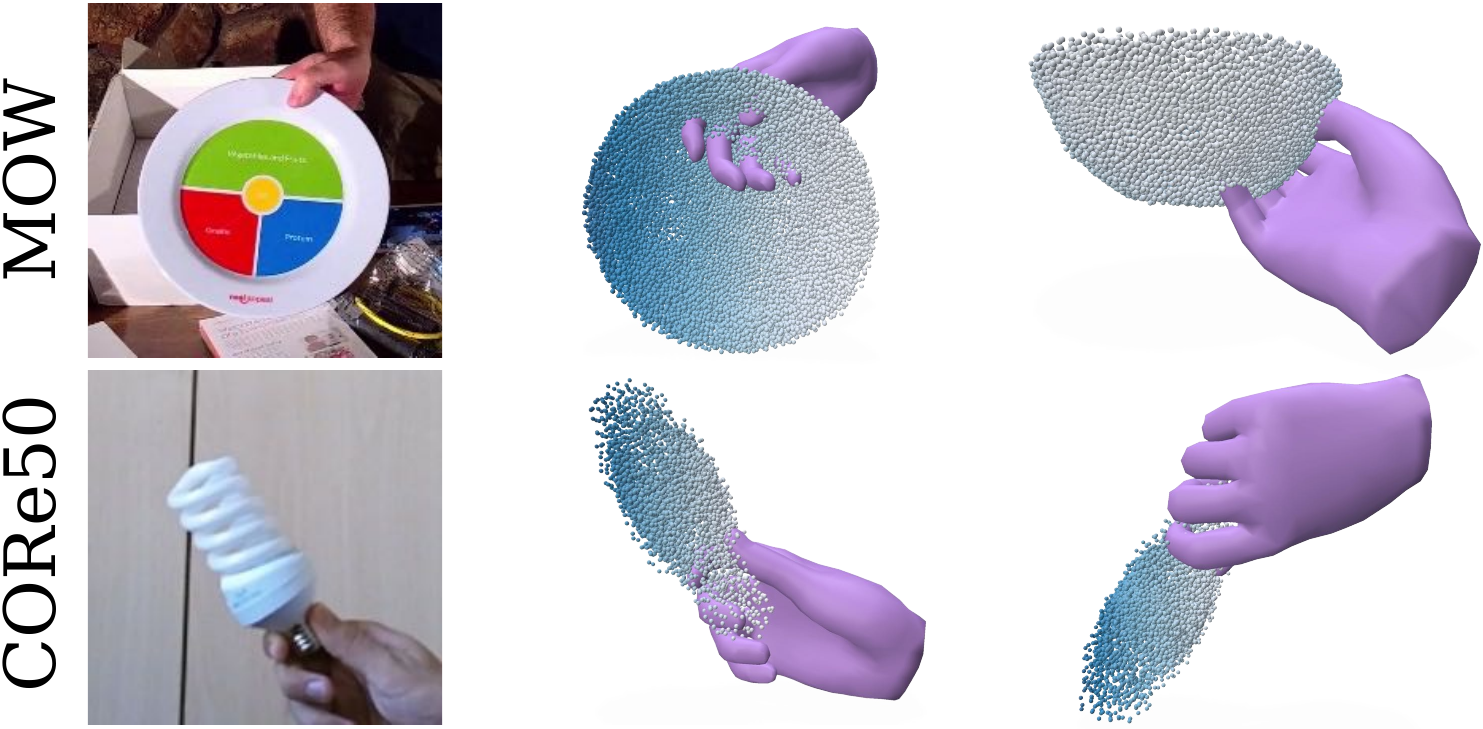}
\vspace{-0.5cm}
\caption{Failure cases on MOW~\cite{cao2021reconstructing} and CORe50~\cite{lomonaco2017core50} images.}
\label{sota:failure}
\vspace{-0.5cm}
\end{figure}

\subsection{Comparison for 3D reconstruction densities}
Figure~\ref{fig:supmat_density} qualitatively compares our reconstructed sparse and dense object point clouds. We observe that our reconstructed sparse and dense point clouds are consistent in the general shape of the object. Our dense point clouds contain more surface details for the manipulated object.

\begin{figure}[t]
  \centering
  \includegraphics[width=1.0\linewidth]{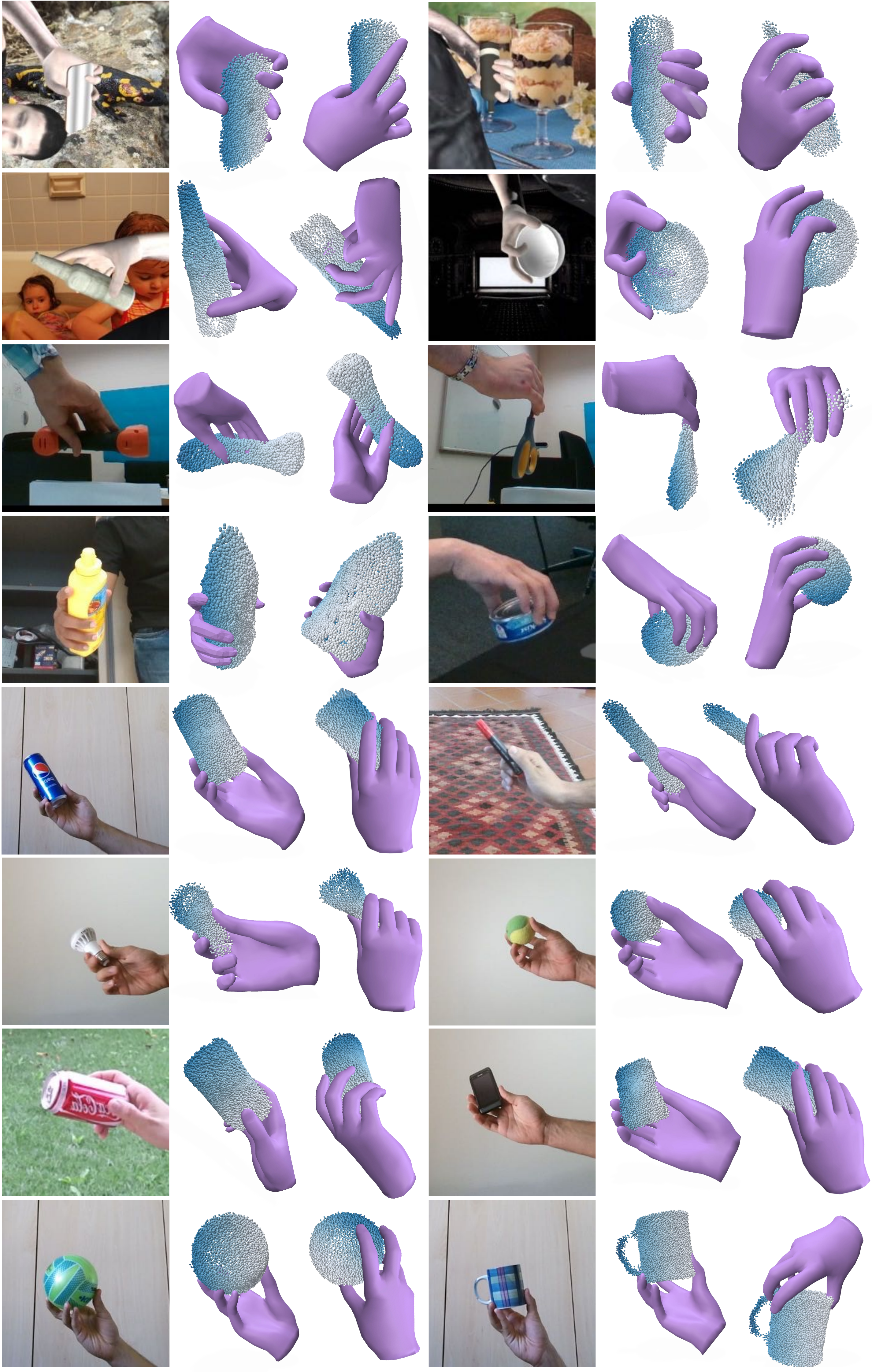}
  \caption{Qualitative results of HORT on synthetic ObMan~\cite{hasson2019learning}, indoor HO3D~\cite{hampali2020honnotate} and DexYCB~\cite{chao2021dexycb}, and in-the-wild CORe50~\cite{lomonaco2017core50} images. Our model shows impressive results on all these domains.}
  \label{fig:supmat_demo}
\end{figure}

\subsection{Impact of imperfect hand poses}
Table~\ref{tab:supmat_T2} ablates the impact of hand accuracy to our model by gradually adding Gaussian noise to ground-truth hand poses. As a result, the hand joint error $\rm E_h$ increases from 7.67$\rm mm$ to 36.11$\rm mm$ with more noises. Our HORT model is robust to noisy hand poses and can still reconstruct plausible 3D hand-held objects.

\subsection{Additional qualitative results}
\begin{figure}[!t]
\vspace{-0.2cm}
  \centering
  \includegraphics[width=1\linewidth]{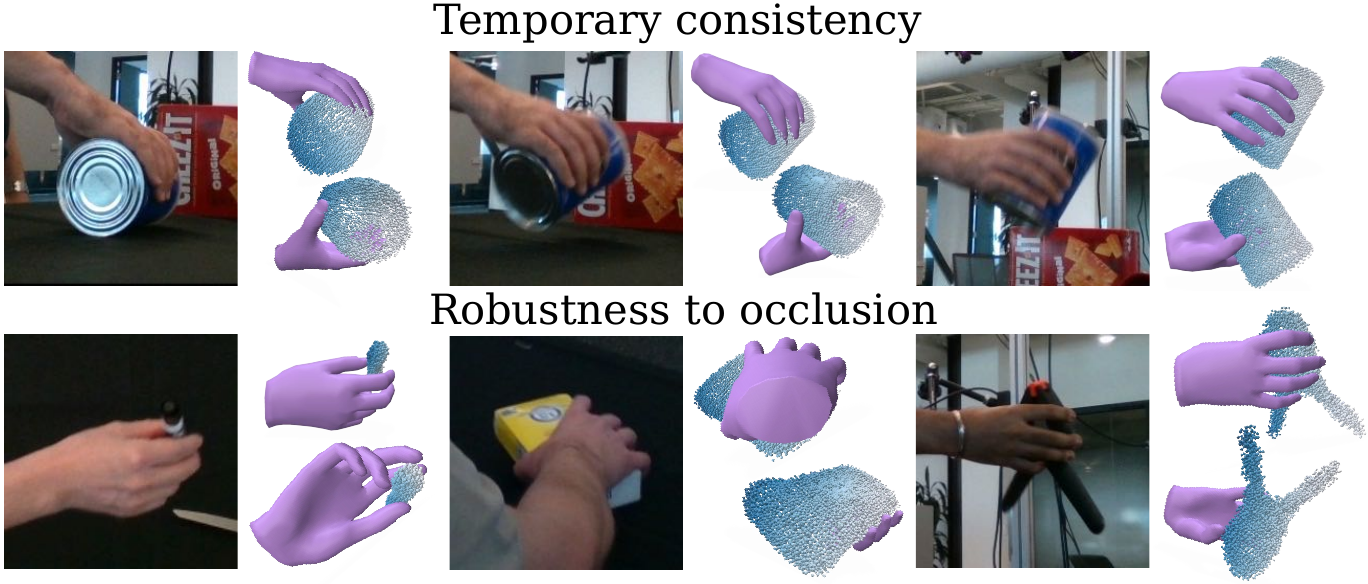}
  \vspace{-0.8cm}
  \caption{Qualitative analysis on DexYCB datasets.}
  \label{fig:rebuttal_temp}
  \vspace{-0.6cm}
\end{figure}

In this section, we present additional qualitative examples in Figure~\ref{fig:supmat_demo} to demonstrate that our approach produces high-quality 3D reconstructions across various challenging scenes. Our model performs well on both the synthetic ObMan~\cite{hasson2019learning} dataset and the real-world HO3D~\cite{hampali2020honnotate} and DexYCB~\cite{chao2021dexycb} datasets. Furthermore, we show that our model can generate reliable predictions on in-the-wild CORe50~\cite{lomonaco2017core50} images, highlighting the ability of our HORT model to effectively generalize to diverse object instances and textures in unconstrained environments.~Figure~\ref{fig:rebuttal_temp} demonstrates that our reconstructed results are temporally consistent and robust to hand occlusion.

\subsection{Failure cases analysis}
While HORT achieves impressive results, it struggles to accurately infer hand-object configurations when the hand is heavily occluded (Figure~\ref{sota:failure}, top) and to recover fine-grained geometry for rare objects (Figure~\ref{sota:failure}, bottom). Scaling training data with a broader range of objects and grasping poses could help mitigate these limitations.

%% file: table/supmat_mesh.tex
\begin{table}[t]
\centering
\footnotesize
\caption{Comparison with previous methods on DexYCB.}
\setlength{\tabcolsep}{1.2pt}
\renewcommand\arraystretch{1.0}
\vspace{-0.2cm}
\begin{tabular}{cccccccc}
\toprule
&Methods & mesh&${\rm {FS}@5}\uparrow$&${\rm {FS}@10}\uparrow$&${\rm CD}\downarrow$&${\rm Cr}\uparrow$&${\rm Pd}\downarrow$ \\ \midrule
R1&GF~\cite{karunratanakul2020grasping} &\checkmark&0.39&0.66&4.5&0.96&0.92 \\
R2&AlignSDF~\cite{chen2022alignsdf}&\checkmark&0.41&0.68&3.9&0.97&1.08 \\
R3&gSDF~\cite{chen2023gsdf}&\checkmark&0.44&0.71&3.4&0.95&0.94 \\
R4&D-SCO~\cite{fu2024d}&$\times$&\textbf{0.63}&0.82&1.3&-&- \\ \midrule
R5&HORT (Ours) &$\times$&\textbf{0.63}&\textbf{0.85}&\textbf{1.1}&\textbf{0.98}&\textbf{0.90} \\ 
R6&HORT (Ours)
&\checkmark&0.62&\textbf{0.85}&\textbf{1.1}&\textbf{0.98}&\textbf{0.90} \\ 
\bottomrule
\end{tabular}
\label{tab:supmat_T1}
\vspace{-0.3cm}
\end{table}

%% file: table/supmat_hand.tex
\begin{table}[!t]
\centering
\footnotesize
\caption{Ablation studies of hand accuracy on DexYCB dataset.}
\vspace{-0.2cm}
\setlength{\tabcolsep}{2.0pt}
\renewcommand\arraystretch{1.0}
\begin{tabular}{ccccccccc}
\toprule
Methods & Noise&$\rm {E_h}\downarrow$&${\rm {FS}@5}\uparrow$&${\rm {FS}@10}\uparrow$&${\rm CD}\downarrow$&${\rm C_{r}}\uparrow$& ${\rm P_{d}}\downarrow$\\ \midrule
HORT (Ours)&$\sigma=0.0$&0.00&\textbf{0.64}&\textbf{0.88}&\textbf{1.0}&\textbf{0.98}&\textbf{0.88}\\
HORT (Ours)&$\sigma=0.1$&7.67&0.60&0.85&1.3&0.96&0.93\\
HORT (Ours)&$\sigma=0.5$&36.11&0.54&0.79&1.6&0.91&1.02\\ 
 \bottomrule
\end{tabular}
\vspace{-0.3cm}
\label{tab:supmat_T2}
\end{table}

%% file: main.bbl
\begin{thebibliography}{102}
\providecommand{\natexlab}[1]{#1}
\providecommand{\url}[1]{\texttt{#1}}
\expandafter\ifx\csname urlstyle\endcsname\relax
  \providecommand{\doi}[1]{doi: #1}\else
  \providecommand{\doi}{doi: \begingroup \urlstyle{rm}\Url}\fi

\bibitem[Aumentado-Armstrong et~al.(2022)Aumentado-Armstrong, Tsogkas, Dickinson, and Jepson]{aumentado2022representing}
Tristan Aumentado-Armstrong, Stavros Tsogkas, Sven Dickinson, and Allan~D Jepson.
\newblock Representing {3D} shapes with probabilistic directed distance fields.
\newblock In \emph{CVPR}, 2022.

\bibitem[Baek et~al.(2019)Baek, Kim, and Kim]{baek2019pushing}
Seungryul Baek, Kwang~In Kim, and Tae-Kyun Kim.
\newblock Pushing the envelope for {RGB}-based dense {3D} hand pose estimation via neural rendering.
\newblock In \emph{CVPR}, 2019.

\bibitem[Ballan et~al.(2012)Ballan, Taneja, Gall, Gool, and Pollefeys]{ballan2012motion}
Luca Ballan, Aparna Taneja, J{\"u}rgen Gall, Luc~Van Gool, and Marc Pollefeys.
\newblock Motion capture of hands in action using discriminative salient points.
\newblock In \emph{ECCV}, 2012.

\bibitem[Banerjee et~al.(2024)Banerjee, Shkodrani, Moulon, Hampali, Zhang, Fountain, Miller, Basol, Newcombe, Wang, et~al.]{banerjee2024introducing}
Prithviraj Banerjee, Sindi Shkodrani, Pierre Moulon, Shreyas Hampali, Fan Zhang, Jade Fountain, Edward Miller, Selen Basol, Richard Newcombe, Robert Wang, et~al.
\newblock Introducing {HOT3D}: An egocentric dataset for {3D} hand and object tracking.
\newblock \emph{arXiv:2406.09598}, 2024.

\bibitem[Boukhayma et~al.(2019)Boukhayma, Bem, and Torr]{boukhayma20193d}
Adnane Boukhayma, Rodrigo~de Bem, and Philip~HS Torr.
\newblock {3D} hand shape and pose from images in the wild.
\newblock In \emph{CVPR}, 2019.

\bibitem[Calli et~al.(2015)Calli, Singh, Walsman, Srinivasa, Abbeel, and Dollar]{calli2015ycb}
Berk Calli, Arjun Singh, Aaron Walsman, Siddhartha Srinivasa, Pieter Abbeel, and Aaron~M Dollar.
\newblock The {YCB} object and model set: Towards common benchmarks for manipulation research.
\newblock In \emph{ICAR}, 2015.

\bibitem[Cao et~al.(2021)Cao, Radosavovic, Kanazawa, and Malik]{cao2021reconstructing}
Zhe Cao, Ilija Radosavovic, Angjoo Kanazawa, and Jitendra Malik.
\newblock Reconstructing hand-object interactions in the wild.
\newblock In \emph{ICCV}, 2021.

\bibitem[Chang et~al.(2015)Chang, Funkhouser, Guibas, Hanrahan, Huang, Li, Savarese, Savva, Song, Su, et~al.]{chang2015shapenet}
Angel~X Chang, Thomas Funkhouser, Leonidas Guibas, Pat Hanrahan, Qixing Huang, Zimo Li, Silvio Savarese, Manolis Savva, Shuran Song, Hao Su, et~al.
\newblock {ShapeNet}: An information-rich {3D} model repository.
\newblock \emph{arXiv preprint arXiv:1512.03012}, 2015.

\bibitem[Chao et~al.(2021)Chao, Yang, Xiang, Molchanov, Handa, Tremblay, Narang, Van~Wyk, Iqbal, Birchfield, et~al.]{chao2021dexycb}
Yu-Wei Chao, Wei Yang, Yu Xiang, Pavlo Molchanov, Ankur Handa, Jonathan Tremblay, Yashraj~S Narang, Karl Van~Wyk, Umar Iqbal, Stan Birchfield, et~al.
\newblock {DexYCB}: A benchmark for capturing hand grasping of objects.
\newblock In \emph{CVPR}, 2021.

\bibitem[Chao et~al.(2022)Chao, Paxton, Xiang, Yang, Sundaralingam, Chen, Murali, Cakmak, and Fox]{chao2022handoversim}
Yu-Wei Chao, Chris Paxton, Yu Xiang, Wei Yang, Balakumar Sundaralingam, Tao Chen, Adithyavairavan Murali, Maya Cakmak, and Dieter Fox.
\newblock {HandoverSim}: A simulation framework and benchmark for human-to-robot object handovers.
\newblock In \emph{ICRA}, 2022.

\bibitem[Chen et~al.(2021)Chen, Liu, Ma, Chang, Wang, Chen, Guo, Wan, and Zheng]{chen2021camera}
Xingyu Chen, Yufeng Liu, Chongyang Ma, Jianlong Chang, Huayan Wang, Tian Chen, Xiaoyan Guo, Pengfei Wan, and Wen Zheng.
\newblock Camera-space hand mesh recovery via semantic aggregation and adaptive {2D-1D} registration.
\newblock In \emph{CVPR}, 2021.

\bibitem[Chen and Zhang(2019)]{chen2019learning}
Zhiqin Chen and Hao Zhang.
\newblock Learning implicit fields for generative shape modeling.
\newblock In \emph{CVPR}, 2019.

\bibitem[Chen et~al.(2020)Chen, Huang, Yu, Xue, Han, Guo, and Wang]{chen2020towards}
Zerui Chen, Yan Huang, Hongyuan Yu, Bin Xue, Ke Han, Yiru Guo, and Liang Wang.
\newblock Towards part-aware monocular {3D} human pose estimation: An architecture search approach.
\newblock In \emph{ECCV}, 2020.

\bibitem[Chen et~al.(2022{\natexlab{a}})Chen, Hasson, Schmid, and Laptev]{chen2022alignsdf}
Zerui Chen, Yana Hasson, Cordelia Schmid, and Ivan Laptev.
\newblock {AlignSDF}: {Pose-Aligned} signed distance fields for hand-object reconstruction.
\newblock In \emph{ECCV}, 2022{\natexlab{a}}.

\bibitem[Chen et~al.(2022{\natexlab{b}})Chen, Huang, Yu, and Wang]{chen2022learning}
Zerui Chen, Yan Huang, Hongyuan Yu, and Liang Wang.
\newblock Learning a robust part-aware monocular 3d human pose estimator via neural architecture search.
\newblock \emph{IJCV}, 2022{\natexlab{b}}.

\bibitem[Chen et~al.(2023)Chen, Chen, Schmid, and Laptev]{chen2023gsdf}
Zerui Chen, Shizhe Chen, Cordelia Schmid, and Ivan Laptev.
\newblock {gSDF}: Geometry-driven signed distance functions for {3D} hand-object reconstruction.
\newblock In \emph{CVPR}, 2023.

\bibitem[Chen et~al.(2025)Chen, Chen, Arlaud, Laptev, and Schmid]{chen2025vividex}
Zerui Chen, Shizhe Chen, Etienne Arlaud, Ivan Laptev, and Cordelia Schmid.
\newblock {ViViDex}: Learning vision-based dexterous manipulation from human videos.
\newblock In \emph{ICRA}, 2025.

\bibitem[Choe et~al.(2022)Choe, Joung, Rameau, Park, and Kweon]{choe2021deep}
Jaesung Choe, Byeongin Joung, Francois Rameau, Jaesik Park, and In~So Kweon.
\newblock Deep point cloud reconstruction.
\newblock In \emph{ICLR}, 2022.

\bibitem[Christen et~al.(2023)Christen, Yang, P{\'e}rez-D’Arpino, Hilliges, Fox, and Chao]{christen2023learning}
Sammy Christen, Wei Yang, Claudia P{\'e}rez-D’Arpino, Otmar Hilliges, Dieter Fox, and Yu-Wei Chao.
\newblock Learning human-to-robot handovers from point clouds.
\newblock In \emph{CVPR}, 2023.

\bibitem[Christen et~al.(2024)Christen, Feng, Yang, Chao, Hilliges, and Song]{christen2024synh2r}
Sammy Christen, Lan Feng, Wei Yang, Yu-Wei Chao, Otmar Hilliges, and Jie Song.
\newblock {SynH2R}: Synthesizing hand-object motions for learning human-to-robot handovers.
\newblock In \emph{ICRA}, 2024.

\bibitem[Damen et~al.(2018)Damen, Doughty, Farinella, Fidler, Furnari, Kazakos, Moltisanti, Munro, Perrett, Price, et~al.]{damen2018scaling}
Dima Damen, Hazel Doughty, Giovanni~Maria Farinella, Sanja Fidler, Antonino Furnari, Evangelos Kazakos, Davide Moltisanti, Jonathan Munro, Toby Perrett, Will Price, et~al.
\newblock Scaling egocentric vision: The {EPIC-KITCHENS} dataset.
\newblock In \emph{ECCV}, 2018.

\bibitem[Edelsbrunner et~al.(1983)Edelsbrunner, Kirkpatrick, and Seidel]{edelsbrunner1983shape}
Herbert Edelsbrunner, David Kirkpatrick, and Raimund Seidel.
\newblock On the shape of a set of points in the plane.
\newblock \emph{IEEE Transactions on information theory}, 1983.

\bibitem[Fan et~al.(2024{\natexlab{a}})Fan, Ohkawa, Yang, Lin, Zhou, Zhou, Liang, Gao, Zhang, Zhang, et~al.]{fan2024benchmarks}
Zicong Fan, Takehiko Ohkawa, Linlin Yang, Nie Lin, Zhishan Zhou, Shihao Zhou, Jiajun Liang, Zhong Gao, Xuanyang Zhang, Xue Zhang, et~al.
\newblock Benchmarks and challenges in pose estimation for egocentric hand interactions with objects.
\newblock In \emph{ECCV}, 2024{\natexlab{a}}.

\bibitem[Fan et~al.(2024{\natexlab{b}})Fan, Parelli, Kadoglou, Chen, Kocabas, Black, and Hilliges]{fan2024hold}
Zicong Fan, Maria Parelli, Maria~Eleni Kadoglou, Xu Chen, Muhammed Kocabas, Michael~J Black, and Otmar Hilliges.
\newblock {HOLD}: Category-agnostic {3D} reconstruction of interacting hands and objects from video.
\newblock In \emph{CVPR}, 2024{\natexlab{b}}.

\bibitem[Fu et~al.(2024)Fu, Wang, Zhang, Di, Huang, Leng, Manhardt, Ji, and Tombari]{fu2024d}
Bowen Fu, Gu Wang, Chenyangguang Zhang, Yan Di, Ziqin Huang, Zhiying Leng, Fabian Manhardt, Xiangyang Ji, and Federico Tombari.
\newblock {D-SCo}: Dual-stream conditional diffusion for monocular hand-held object reconstruction.
\newblock In \emph{ECCV}, 2024.

\bibitem[Grauman et~al.(2022)Grauman, Westbury, Byrne, Chavis, Furnari, Girdhar, Hamburger, Jiang, Liu, Liu, et~al.]{grauman2022ego4d}
Kristen Grauman, Andrew Westbury, Eugene Byrne, Zachary Chavis, Antonino Furnari, Rohit Girdhar, Jackson Hamburger, Hao Jiang, Miao Liu, Xingyu Liu, et~al.
\newblock {Ego4D}: Around the world in 3,000 hours of egocentric video.
\newblock In \emph{CVPR}, 2022.

\bibitem[Groueix et~al.(2018)Groueix, Fisher, Kim, Russell, and Aubry]{groueix2018papier}
Thibault Groueix, Matthew Fisher, Vladimir~G Kim, Bryan~C Russell, and Mathieu Aubry.
\newblock A papier-m{\^a}ch{\'e} approach to learning {3D} surface generation.
\newblock In \emph{CVPR}, 2018.

\bibitem[Hampali et~al.(2020)Hampali, Rad, Oberweger, and Lepetit]{hampali2020honnotate}
Shreyas Hampali, Mahdi Rad, Markus Oberweger, and Vincent Lepetit.
\newblock {HO}nnotate: A method for 3{D} annotation of hand and object poses.
\newblock In \emph{CVPR}, 2020.

\bibitem[Hampali et~al.(2021)Hampali, Sarkar, and Lepetit]{hampali2021ho}
Shreyas Hampali, Sayan~Deb Sarkar, and Vincent Lepetit.
\newblock {HO3Dv3}: Improving the accuracy of hand-object annotations of the {HO-3D} dataset.
\newblock \emph{arXiv preprint arXiv:2107.00887}, 2021.

\bibitem[Hampali et~al.(2022)Hampali, Sarkar, Rad, and Lepetit]{hampali2022keypoint}
Shreyas Hampali, Sayan~Deb Sarkar, Mahdi Rad, and Vincent Lepetit.
\newblock {Keypoint Transformer}: Solving joint identification in challenging hands and object interactions for accurate {3D} pose estimation.
\newblock In \emph{CVPR}, 2022.

\bibitem[Hampali et~al.(2023)Hampali, Hodan, Tran, Ma, Keskin, and Lepetit]{hampali2023hand}
Shreyas Hampali, Tomas Hodan, Luan Tran, Lingni Ma, Cem Keskin, and Vincent Lepetit.
\newblock {In-Hand} {3D} object scanning from an {RGB} sequence.
\newblock In \emph{CVPR}, 2023.

\bibitem[Hasson et~al.(2019)Hasson, Varol, Tzionas, Kalevatykh, Black, Laptev, and Schmid]{hasson2019learning}
Yana Hasson, Gul Varol, Dimitrios Tzionas, Igor Kalevatykh, Michael~J Black, Ivan Laptev, and Cordelia Schmid.
\newblock Learning joint reconstruction of hands and manipulated objects.
\newblock In \emph{CVPR}, 2019.

\bibitem[Hasson et~al.(2021)Hasson, Varol, Schmid, and Laptev]{hasson2021towards}
Yana Hasson, G{\"u}l Varol, Cordelia Schmid, and Ivan Laptev.
\newblock Towards unconstrained joint hand-object reconstruction from {RGB} videos.
\newblock In \emph{3DV}, 2021.

\bibitem[Heap and Hogg(1996)]{heap1996towards}
Tony Heap and David Hogg.
\newblock Towards {3D} hand tracking using a deformable model.
\newblock In \emph{FG}, 1996.

\bibitem[Hu et~al.(2024)Hu, Zhang, Chen, Li, Wang, Liu, and Sun]{hu2024learning}
Junxing Hu, Hongwen Zhang, Zerui Chen, Mengcheng Li, Yunlong Wang, Yebin Liu, and Zhenan Sun.
\newblock Learning explicit contact for implicit reconstruction of hand-held objects from monocular images.
\newblock In \emph{AAAI}, 2024.

\bibitem[Huang et~al.(2022)Huang, Ji, He, Sun, He, Shuai, Ouyang, and Zhou]{huang2022reconstructing}
Di Huang, Xiaopeng Ji, Xingyi He, Jiaming Sun, Tong He, Qing Shuai, Wanli Ouyang, and Xiaowei Zhou.
\newblock Reconstructing hand-held objects from monocular video.
\newblock In \emph{SIGGRAPH Asia}, 2022.

\bibitem[Iqbal et~al.(2018)Iqbal, Molchanov, Gall, and Kautz]{iqbal2018hand}
Umar Iqbal, Pavlo Molchanov, Thomas Breuel~Juergen Gall, and Jan Kautz.
\newblock Hand pose estimation via latent 2.5{D} heatmap regression.
\newblock In \emph{ECCV}, 2018.

\bibitem[Karunratanakul et~al.(2020)Karunratanakul, Yang, Zhang, Black, Muandet, and Tang]{karunratanakul2020grasping}
Korrawe Karunratanakul, Jinlong Yang, Yan Zhang, Michael~J Black, Krikamol Muandet, and Siyu Tang.
\newblock {Grasping Field}: Learning implicit representations for human grasps.
\newblock In \emph{3DV}, 2020.

\bibitem[Kazhdan et~al.(2006)Kazhdan, Bolitho, and Hoppe]{kazhdan2006poisson}
Michael Kazhdan, Matthew Bolitho, and Hugues Hoppe.
\newblock Poisson surface reconstruction.
\newblock In \emph{SGP}, 2006.

\bibitem[Kim and Kim(2024)]{kim2024multi}
Donghwan Kim and Tae-Kyun Kim.
\newblock Multi-hypotheses conditioned point cloud diffusion for 3d human reconstruction from occluded images.
\newblock \emph{NeurIPS}, 2024.

\bibitem[Kingma and Ba(2014)]{kingma2014adam}
Diederik~P Kingma and Jimmy Ba.
\newblock Adam: A method for stochastic optimization.
\newblock \emph{arXiv preprint arXiv:1412.6980}, 2014.

\bibitem[Kulon et~al.(2019)Kulon, Wang, G{\"{u}}ler, Bronstein, and Zafeiriou]{kulon2019rec}
Dominik Kulon, Haoyang Wang, Riza~Alp G{\"{u}}ler, Michael~M. Bronstein, and Stefanos Zafeiriou.
\newblock Single image {3D} hand reconstruction with mesh convolutions.
\newblock In \emph{BMVC}, 2019.

\bibitem[Kulon et~al.(2020)Kulon, G{\"u}ler, Kokkinos, Bronstein, and Zafeiriou]{Kulon2020weaklysupervisedmh}
Dominik Kulon, Riza~Alp G{\"u}ler, I. Kokkinos, M. Bronstein, and S. Zafeiriou.
\newblock Weakly-supervised mesh-convolutional hand reconstruction in the wild.
\newblock In \emph{CVPR}, 2020.

\bibitem[Lepetit(2020)]{lepetit2020recent_pose_advances}
Vincent Lepetit.
\newblock Recent advances in {3D} object and hand pose estimation.
\newblock \emph{arXiv preprint arXiv:2006.05927}, 2020.

\bibitem[Li et~al.(2023)Li, Yang, Zhen, Lin, Zhan, Zhong, Xu, Wu, and Lu]{li2023chord}
Kailin Li, Lixin Yang, Haoyu Zhen, Zenan Lin, Xinyu Zhan, Licheng Zhong, Jian Xu, Kejian Wu, and Cewu Lu.
\newblock {CHORD}: Category-level hand-held object reconstruction via shape deformation.
\newblock In \emph{ICCV}, 2023.

\bibitem[Li et~al.(2022)Li, An, Zhang, Wu, Chen, Yu, and Liu]{li2022interacting}
Mengcheng Li, Liang An, Hongwen Zhang, Lianpeng Wu, Feng Chen, Tao Yu, and Yebin Liu.
\newblock Interacting attention graph for single image two-hand reconstruction.
\newblock In \emph{CVPR}, 2022.

\bibitem[Lin et~al.(2021)Lin, Wang, and Liu]{lin2021end}
Kevin Lin, Lijuan Wang, and Zicheng Liu.
\newblock End-to-end human pose and mesh reconstruction with transformers.
\newblock In \emph{CVPR}, 2021.

\bibitem[Liu et~al.(2022)Liu, Liu, Jiang, Lyu, Wan, Shen, Liang, Fu, Wang, and Yi]{liu2022hoi4d}
Yunze Liu, Yun Liu, Che Jiang, Kangbo Lyu, Weikang Wan, Hao Shen, Boqiang Liang, Zhoujie Fu, He Wang, and Li Yi.
\newblock {HOI4D}: A {4D} egocentric dataset for category-level human-object interaction.
\newblock In \emph{CVPR}, 2022.

\bibitem[Liu et~al.(2025)Liu, Long, Yang, Liu, Habermann, Theobalt, Ma, and Wang]{liu2025easyhoi}
Yumeng Liu, Xiaoxiao Long, Zemin Yang, Yuan Liu, Marc Habermann, Christian Theobalt, Yuexin Ma, and Wenping Wang.
\newblock {EasyHOI}: Unleashing the power of large models for reconstructing hand-object interactions in the wild.
\newblock In \emph{CVPR}, 2025.

\bibitem[Lomonaco and Maltoni(2017)]{lomonaco2017core50}
Vincenzo Lomonaco and Davide Maltoni.
\newblock {CORe50}: a new dataset and benchmark for continuous object recognition.
\newblock In \emph{CoRL}, 2017.

\bibitem[Lorensen and Cline(1987)]{lorensen1987marching}
William~E Lorensen and Harvey~E Cline.
\newblock {Marching Cubes}: A high resolution {3D} surface construction algorithm.
\newblock \emph{TOG}, 1987.

\bibitem[Lv et~al.(2021)Lv, Xu, Yang, Qian, Mao, and Lu]{lv2021handtailor}
Jun Lv, Wenqiang Xu, Lixin Yang, Sucheng Qian, Chongzhao Mao, and Cewu Lu.
\newblock {HandTailor}: Towards high-precision monocular {3D} hand recovery.
\newblock In \emph{BMVC}, 2021.

\bibitem[Mandikal and Grauman(2022)]{mandikal2022dexvip}
Priyanka Mandikal and Kristen Grauman.
\newblock {DexVIP}: Learning dexterous grasping with human hand pose priors from video.
\newblock In \emph{CoRL}, 2022.

\bibitem[Mehta et~al.(2017)Mehta, Rhodin, Casas, Fua, Sotnychenko, Xu, and Theobalt]{mehta2017monocular}
Dushyant Mehta, Helge Rhodin, Dan Casas, Pascal Fua, Oleksandr Sotnychenko, Weipeng Xu, and Christian Theobalt.
\newblock Monocular {3D} human pose estimation in the wild using improved {CNN} supervision.
\newblock In \emph{3DV}, 2017.

\bibitem[Melas-Kyriazi et~al.(2023)Melas-Kyriazi, Rupprecht, and Vedaldi]{melas2023pc2}
Luke Melas-Kyriazi, Christian Rupprecht, and Andrea Vedaldi.
\newblock {PC2}: Projection-conditioned point cloud diffusion for single-image 3d reconstruction.
\newblock In \emph{CVPR}, 2023.

\bibitem[Meng et~al.(2022)Meng, Jin, Liu, Qian, Lin, Ouyang, and Luo]{meng20223d}
Hao Meng, Sheng Jin, Wentao Liu, Chen Qian, Mengxiang Lin, Wanli Ouyang, and Ping Luo.
\newblock {3D} interacting hand pose estimation by hand de-occlusion and removal.
\newblock In \emph{ECCV}, 2022.

\bibitem[Mescheder et~al.(2019)Mescheder, Oechsle, Niemeyer, Nowozin, and Geiger]{mescheder2019occupancy}
Lars Mescheder, Michael Oechsle, Michael Niemeyer, Sebastian Nowozin, and Andreas Geiger.
\newblock Occupancy {Networks}: Learning {3D} reconstruction in function space.
\newblock In \emph{CVPR}, 2019.

\bibitem[Mildenhall et~al.(2020)Mildenhall, Srinivasan, Tancik, Barron, Ramamoorthi, and Ng]{mildenhall2020nerf}
B Mildenhall, PP Srinivasan, M Tancik, JT Barron, R Ramamoorthi, and R Ng.
\newblock {NeRF}: Representing scenes as neural radiance fields for view synthesis.
\newblock In \emph{ECCV}, 2020.

\bibitem[Miller and Allen(2004)]{miller2004graspit}
Andrew~T Miller and Peter~K Allen.
\newblock {Graspit!} a versatile simulator for robotic grasping.
\newblock \emph{RAM}, 2004.

\bibitem[Moon et~al.(2018)Moon, Chang, and Lee]{moon2018v2v}
Gyeongsik Moon, Ju~Yong Chang, and Kyoung~Mu Lee.
\newblock {V2V-PoseNet}: Voxel-to-voxel prediction network for accurate {3D} hand and human pose estimation from a single depth map.
\newblock In \emph{CVPR}, 2018.

\bibitem[Mueller et~al.(2018)Mueller, Bernard, Sotnychenko, Mehta, Sridhar, Casas, and Theobalt]{mueller2018ganerated}
Franziska Mueller, Florian Bernard, Oleksandr Sotnychenko, Dushyant Mehta, Srinath Sridhar, Dan Casas, and Christian Theobalt.
\newblock Ganerated hands for real-time {3D} hand tracking from monocular {RGB}.
\newblock In \emph{CVPR}, 2018.

\bibitem[Mueller et~al.(2019)Mueller, Davis, Bernard, Sotnychenko, Verschoor, Otaduy, Casas, and Theobalt]{mueller2019real}
Franziska Mueller, Micah Davis, Florian Bernard, Oleksandr Sotnychenko, Mickeal Verschoor, Miguel~A Otaduy, Dan Casas, and Christian Theobalt.
\newblock Real-time pose and shape reconstruction of two interacting hands with a single depth camera.
\newblock \emph{TOG}, 2019.

\bibitem[Nichol et~al.(2022)Nichol, Jun, Dhariwal, Mishkin, and Chen]{nichol2022point}
Alex Nichol, Heewoo Jun, Prafulla Dhariwal, Pamela Mishkin, and Mark Chen.
\newblock {Point-E}: A system for generating {3D} point clouds from complex prompts.
\newblock \emph{arXiv preprint arXiv:2212.08751}, 2022.

\bibitem[Oikonomidis et~al.(2011)Oikonomidis, Kyriazis, and Argyros]{oikonomidis2011full}
Iason Oikonomidis, Nikolaos Kyriazis, and Antonis~A Argyros.
\newblock Full {DOF} tracking of a hand interacting with an object by modeling occlusions and physical constraints.
\newblock In \emph{ICCV}, 2011.

\bibitem[Oquab et~al.(2023)Oquab, Darcet, Moutakanni, Vo, Szafraniec, Khalidov, Fernandez, Haziza, Massa, El-Nouby, et~al.]{oquab2023dinov2}
Maxime Oquab, Timoth{\'e}e Darcet, Th{\'e}o Moutakanni, Huy Vo, Marc Szafraniec, Vasil Khalidov, Pierre Fernandez, Daniel Haziza, Francisco Massa, Alaaeldin El-Nouby, et~al.
\newblock {DINOv2}: Learning robust visual features without supervision.
\newblock \emph{arXiv:2304.07193}, 2023.

\bibitem[Park et~al.(2019)Park, Florence, Straub, Newcombe, and Lovegrove]{park2019deepsdf}
Jeong~Joon Park, Peter Florence, Julian Straub, Richard Newcombe, and Steven Lovegrove.
\newblock {DeepSDF}: Learning continuous signed distance functions for shape representation.
\newblock In \emph{CVPR}, 2019.

\bibitem[Pavlakos et~al.(2024)Pavlakos, Shan, Radosavovic, Kanazawa, Fouhey, and Malik]{pavlakos2024reconstructing}
Georgios Pavlakos, Dandan Shan, Ilija Radosavovic, Angjoo Kanazawa, David Fouhey, and Jitendra Malik.
\newblock Reconstructing hands in 3d with transformers.
\newblock In \emph{CVPR}, 2024.

\bibitem[Potamias et~al.(2025)Potamias, Zhang, Deng, and Zafeiriou]{potamias2024wilor}
Rolandos~Alexandros Potamias, Jinglei Zhang, Jiankang Deng, and Stefanos Zafeiriou.
\newblock {WiLoR}: End-to-end {3D} hand localization and reconstruction in-the-wild.
\newblock In \emph{CVPR}, 2025.

\bibitem[Prakash et~al.(2024)Prakash, Chang, et~al.]{prakash2023learning}
Aditya Prakash, Matthew Chang, et~al.
\newblock 3d reconstruction of objects in hands without real world 3d supervision.
\newblock In \emph{ECCV}, 2024.

\bibitem[Qi et~al.(2017)Qi, Su, Mo, and Guibas]{qi2017pointnet}
Charles~R Qi, Hao Su, Kaichun Mo, and Leonidas~J Guibas.
\newblock {PointNet}: Deep learning on point sets for {3D} classification and segmentation.
\newblock In \emph{CVPR}, 2017.

\bibitem[Qi et~al.(2024)Qi, Zhao, Salzmann, and Mathis]{qi2024hoisdf}
Haozhe Qi, Chen Zhao, Mathieu Salzmann, and Alexander Mathis.
\newblock {HOISDF}: Constraining {3D} hand-object pose estimation with global signed distance fields.
\newblock In \emph{CVPR}, 2024.

\bibitem[Qin et~al.(2022)Qin, Wu, Liu, Jiang, Yang, Fu, and Wang]{qin2022dexmv}
Yuzhe Qin, Yueh-Hua Wu, Shaowei Liu, Hanwen Jiang, Ruihan Yang, Yang Fu, and Xiaolong Wang.
\newblock {DexMV}: Imitation learning for dexterous manipulation from human videos.
\newblock In \emph{ECCV}, 2022.

\bibitem[Rehg and Kanade(1994)]{rehg1994visual}
James~M Rehg and Takeo Kanade.
\newblock Visual tracking of high {DOF} articulated structures: an application to human hand tracking.
\newblock In \emph{ECCV}, 1994.

\bibitem[Romero et~al.(2017)Romero, Tzionas, and Black]{MANO:SIGGRAPHASIA:2017}
Javier Romero, Dimitrios Tzionas, and Michael~J. Black.
\newblock Embodied {Hands}: Modeling and capturing hands and bodies together.
\newblock \emph{TOG}, 2017.

\bibitem[Rong et~al.(2021)Rong, Shiratori, and Joo]{rong2021frankmocap}
Yu Rong, Takaaki Shiratori, and Hanbyul Joo.
\newblock {FrankMocap}: A monocular {3D} whole-body pose estimation system via regression and integration.
\newblock In \emph{ICCV}, 2021.

\bibitem[Shan et~al.(2020)Shan, Geng, Shu, and Fouhey]{shan2020understanding}
Dandan Shan, Jiaqi Geng, Michelle Shu, and David~F Fouhey.
\newblock Understanding human hands in contact at internet scale.
\newblock In \emph{CVPR}, 2020.

\bibitem[Song et~al.(2021)Song, Sohl-Dickstein, Kingma, Kumar, Ermon, and Poole]{song2020score}
Yang Song, Jascha Sohl-Dickstein, Diederik~P Kingma, Abhishek Kumar, Stefano Ermon, and Ben Poole.
\newblock Score-based generative modeling through stochastic differential equations.
\newblock In \emph{ICLR}, 2021.

\bibitem[Spurr et~al.(2021)Spurr, Dahiya, Wang, Zhang, and Hilliges]{spurr2021self}
Adrian Spurr, Aneesh Dahiya, Xi Wang, Xucong Zhang, and Otmar Hilliges.
\newblock Self-supervised {3D} hand pose estimation from monocular {RGB} via contrastive learning.
\newblock In \emph{ICCV}, 2021.

\bibitem[Sun et~al.(2018)Sun, Xiao, Wei, Liang, and Wei]{sun2018integral}
Xiao Sun, Bin Xiao, Fangyin Wei, Shuang Liang, and Yichen Wei.
\newblock Integral human pose regression.
\newblock In \emph{ECCV}, 2018.

\bibitem[Tang et~al.(2014)Tang, Jin~Chang, Tejani, and Kim]{tang2014latent}
Danhang Tang, Hyung Jin~Chang, Alykhan Tejani, and Tae-Kyun Kim.
\newblock Latent regression forest: Structured estimation of 3{D} articulated hand posture.
\newblock In \emph{CVPR}, 2014.

\bibitem[Tatarchenko et~al.(2019)Tatarchenko, Richter, Ranftl, Li, Koltun, and Brox]{tatarchenko2019single}
Maxim Tatarchenko, Stephan~R Richter, Ren{\'e} Ranftl, Zhuwen Li, Vladlen Koltun, and Thomas Brox.
\newblock What do single-view {3D} reconstruction networks learn?
\newblock In \emph{CVPR}, 2019.

\bibitem[Tekin et~al.(2019)Tekin, Bogo, and Pollefeys]{tekin2019h+}
Bugra Tekin, Federica Bogo, and Marc Pollefeys.
\newblock {H+O}: Unified egocentric recognition of 3{D} hand-object poses and interactions.
\newblock In \emph{CVPR}, 2019.

\bibitem[Tse et~al.(2022)Tse, Kim, Leonardis, and Chang]{tse2022collaborative}
Tze Ho~Elden Tse, Kwang~In Kim, Ales Leonardis, and Hyung~Jin Chang.
\newblock Collaborative learning for hand and object reconstruction with attention-guided graph convolution.
\newblock In \emph{CVPR}, 2022.

\bibitem[Vaswani et~al.(2017)Vaswani, Shazeer, Parmar, Uszkoreit, Jones, Gomez, Kaiser, and Polosukhin]{vaswani2017attention}
Ashish Vaswani, Noam Shazeer, Niki Parmar, Jakob Uszkoreit, Llion Jones, Aidan~N Gomez, {\L}ukasz Kaiser, and Illia Polosukhin.
\newblock Attention is all you need.
\newblock In \emph{NeurIPS}, 2017.

\bibitem[Wang et~al.(2020)Wang, Mueller, Bernard, Sorli, Sotnychenko, Qian, Otaduy, Casas, and Theobalt]{wang2020rgb2hands}
Jiayi Wang, Franziska Mueller, Florian Bernard, Suzanne Sorli, Oleksandr Sotnychenko, Neng Qian, Miguel~A Otaduy, Dan Casas, and Christian Theobalt.
\newblock {RGB2Hands}: Real-time tracking of {3D} hand interactions from monocular {RGB} video.
\newblock \emph{TOG}, 2020.

\bibitem[Wang et~al.(2024)Wang, Zhang, Chao, Wen, Guo, and Xiang]{wang2024ho}
Jikai Wang, Qifan Zhang, Yu-Wei Chao, Bowen Wen, Xiaohu Guo, and Yu Xiang.
\newblock {HO-Cap}: A capture system and dataset for {3D} reconstruction and pose tracking of hand-object interaction.
\newblock \emph{arXiv:2406.06843}, 2024.

\bibitem[Wang et~al.(2013)Wang, Min, Zhang, Liu, Xu, Dai, and Chai]{wang2013video}
Yangang Wang, Jianyuan Min, Jianjie Zhang, Yebin Liu, Feng Xu, Qionghai Dai, and Jinxiang Chai.
\newblock Video-based hand manipulation capture through composite motion control.
\newblock \emph{TOG}, 2013.

\bibitem[Wen et~al.(2023)Wen, Tremblay, Blukis, Tyree, M{\"u}ller, Evans, Fox, Kautz, and Birchfield]{wen2023bundlesdf}
Bowen Wen, Jonathan Tremblay, Valts Blukis, Stephen Tyree, Thomas M{\"u}ller, Alex Evans, Dieter Fox, Jan Kautz, and Stan Birchfield.
\newblock {BundleSDF}: Neural 6-{DoF} tracking and {3D} reconstruction of unknown objects.
\newblock In \emph{CVPR}, 2023.

\bibitem[Wu et~al.(2024)Wu, Pavlakos, Gkioxari, and Malik]{wu2024reconstructing}
Jane Wu, Georgios Pavlakos, Georgia Gkioxari, and Jitendra Malik.
\newblock Reconstructing hand-held objects in {3D}.
\newblock \emph{arXiv:2404.06507}, 2024.

\bibitem[Xiang et~al.(2021)Xiang, Wen, Liu, Cao, Wan, Zheng, and Han]{xiang2021snowflakenet}
Peng Xiang, Xin Wen, Yu-Shen Liu, Yan-Pei Cao, Pengfei Wan, Wen Zheng, and Zhizhong Han.
\newblock {SnowflakeNet}: Point cloud completion by snowflake point deconvolution with skip-transformer.
\newblock In \emph{ICCV}, 2021.

\bibitem[Xiong et~al.(2019)Xiong, Zhang, Xiao, Cao, Yu, Zhou, and Yuan]{xiong2019a2j}
Fu Xiong, Boshen Zhang, Yang Xiao, Zhiguo Cao, Taidong Yu, Joey~Tianyi Zhou, and Junsong Yuan.
\newblock {A2J}: Anchor-to-joint regression network for {3D} articulated pose estimation from a single depth image.
\newblock In \emph{ICCV}, 2019.

\bibitem[Yang et~al.(2021)Yang, Zhan, Li, Xu, Li, and Lu]{yang2021cpf}
Lixin Yang, Xinyu Zhan, Kailin Li, Wenqiang Xu, Jiefeng Li, and Cewu Lu.
\newblock {CPF}: Learning a contact potential field to model the hand-object interaction.
\newblock In \emph{ICCV}, 2021.

\bibitem[Yang et~al.(2022{\natexlab{a}})Yang, Li, Zhan, Lv, Xu, Li, and Lu]{yang2022artiboost}
Lixin Yang, Kailin Li, Xinyu Zhan, Jun Lv, Wenqiang Xu, Jiefeng Li, and Cewu Lu.
\newblock {ArtiBoost}: Boosting articulated {3D} hand-object pose estimation via online exploration and synthesis.
\newblock In \emph{CVPR}, 2022{\natexlab{a}}.

\bibitem[Yang et~al.(2022{\natexlab{b}})Yang, Li, Zhan, Wu, Xu, Liu, and Lu]{yang2022oakink}
Lixin Yang, Kailin Li, Xinyu Zhan, Fei Wu, Anran Xu, Liu Liu, and Cewu Lu.
\newblock {OakInk}: A large-scale knowledge repository for understanding hand-object interaction.
\newblock In \emph{CVPR}, 2022{\natexlab{b}}.

\bibitem[Ye et~al.(2022)Ye, Gupta, and Tulsiani]{ye2022s}
Yufei Ye, Abhinav Gupta, and Shubham Tulsiani.
\newblock What's in your hands? {3D} reconstruction of generic objects in hands.
\newblock In \emph{CVPR}, 2022.

\bibitem[Ye et~al.(2023)Ye, Hebbar, Gupta, and Tulsiani]{ye2023diffusion}
Yufei Ye, Poorvi Hebbar, Abhinav Gupta, and Shubham Tulsiani.
\newblock Diffusion-guided reconstruction of everyday hand-object interaction clips.
\newblock In \emph{ICCV}, 2023.

\bibitem[Yu et~al.(2021)Yu, Salzmann, Fua, and Rhodin]{yu2021pcls}
Frank Yu, Mathieu Salzmann, Pascal Fua, and Helge Rhodin.
\newblock {PCLs}: Geometry-aware neural reconstruction of {3D} pose with perspective crop layers.
\newblock In \emph{CVPR}, 2021.

\bibitem[Yuan et~al.(2018)Yuan, Garcia-Hernando, Stenger, Moon, Chang, Lee, Molchanov, Kautz, Honari, Ge, Yuan, Chen, Wang, Yang, Akiyama, Wu, Wan, Madadi, Escalera, Li, Lee, Oikonomidis, Argyros, and Kim]{Yuan_2018_CVPR}
Shanxin Yuan, Guillermo Garcia-Hernando, Björn Stenger, Gyeongsik Moon, Ju~Yong Chang, Kyoung~Mu Lee, Pavlo Molchanov, Jan Kautz, Sina Honari, Liuhao Ge, Junsong Yuan, Xinghao Chen, Guijin Wang, Fan Yang, Kai Akiyama, Yang Wu, Qingfu Wan, Meysam Madadi, Sergio Escalera, Shile Li, Dongheui Lee, Iason Oikonomidis, Antonis Argyros, and Tae-Kyun Kim.
\newblock Depth-based {3D} hand pose estimation: From current achievements to future goals.
\newblock In \emph{CVPR}, 2018.

\bibitem[Zhang et~al.(2024)Zhang, Di, Zhang, Zhai, Manhardt, Tombari, and Ji]{zhang2024ddf}
Chenyangguang Zhang, Yan Di, Ruida Zhang, Guangyao Zhai, Fabian Manhardt, Federico Tombari, and Xiangyang Ji.
\newblock {DDF-HO}: hand-held object reconstruction via conditional directed distance field.
\newblock In \emph{NeurIPS}, 2024.

\bibitem[Zhang et~al.(2025)Zhang, Deng, Ma, and Potamias]{zhang2025hawor}
Jinglei Zhang, Jiankang Deng, Chao Ma, and Rolandos~Alexandros Potamias.
\newblock {HaWoR}: World-space hand motion reconstruction from egocentric videos.
\newblock In \emph{CVPR}, 2025.

\bibitem[Zhu and Damen(2023)]{zhu2023get}
Zhifan Zhu and Dima Damen.
\newblock Get a grip: Reconstructing hand-object stable grasps in egocentric videos.
\newblock \emph{arXiv:2312.15719}, 2023.

\bibitem[Zimmermann and Brox(2017)]{zimmermann2017learning}
Christian Zimmermann and Thomas Brox.
\newblock Learning to estimate {3D} hand pose from single {RGB} images.
\newblock In \emph{ICCV}, 2017.

\end{thebibliography}
